# SAM-based instance segmentation models for the automation of structural damage detection

Zehao Ye, Lucy Lovell, Asaad Faramarzi and Jelena Ninić[*]

*University of Birmingham, United Kingdom*

**Abstract**

Automating visual inspection for capturing defects based on civil structures appearance is crucial due to its currently labour-intensive and time-consuming nature. An important aspect of automated inspection is image acquisition, which is rapid and cost-effective considering the pervasive developments in both software and hardware computing in recent years. Previous studies largely focused on concrete and asphalt, with limited attention to masonry cracks and a lack of publicly available datasets. In this paper, we address these gaps by introducing the "MCrack1300" dataset, consisting of 1,300 annotated images (640 pixels × 640 pixels), covering bricks, broken bricks, and cracks, for instance segmentation. We evaluate several leading algorithms for benchmarking, and propose two novel, automatically executable methods based on the latest visual large-scale model, the prompt-based Segment Anything Model (SAM). We fine-tune the encoder of SAM using Low-Rank Adaptation (LoRA). The first method involves abandoning the prompt encoder and connecting the SAM's encoder to other decoders, while the second method introduces a learnable self-generating prompter. We redesign the feature extractor for seamless integration of the two proposed methods with SAM's encoder. Both proposed methods exceed state-of-the-art performance, surpassing the best benchmark by approximately 3% for all classes and around 6% for cracks specifically. Based on successful detection, we then propose a method based on a monocular camera and the Hough Line Transform to automatically convert images into orthographic projection maps. By incorporating known real sizes of brick units, we accurately estimate crack dimensions, with the results differing by less than 10% from those obtained by laser scanning. Overall, we address important research gaps in masonry crack detection and size estimation by introducing a new dataset, as well as novel SAM-based detection algorithms and monocular photogrammetric methodology, ultimately, offering reliable automated solutions.

[*]*Corresponding author:* Jelena Ninić (j.ninic@bham.ac.uk)

Zehao Ye (zxy239@student.bham.ac.uk)

# 1. Introduction

Civil engineering structures often endure significant physical strains, which can be triggered by natural disasters like earthquakes, catastrophic events such as explosions, or regular wear and tear due to everyday use. These events can result in two possible outcomes: either a total structural collapse or the development of physical damage, often visible in the form of defects such as cracks and spalling (Munawar et al., 2021). Automated reality capture technologies, including camera-based, infrared-based, ultrasonic image-based, and laser image-based methods (Mohan and Poobal, 2018), offer a promising avenue for defect monitoring. Unlike labour-intensive manual inspections reliant on subjective expertise, automated detection ensures increased reliability, objectivity, and reproducibility in quantitative analysis (Fang et al., 2020; McRobbie et al., 2015). The growing interest in automated detection of defects such as cracks aligns with the rapid and cost-effective nature of modern image acquisition, offering an efficient alternative to manual inspection.

Masonry is a commonly used construction material known for its durability and strength, which is easily obtained from local sources worldwide. As a result, numerous historical and cultural heritage buildings, including long-standing structures such as Victorian-era railway tunnels that have endured for over 150 years, are constructed using masonry techniques in the UK (Atkinson et al., 2021; Seo et al., 2023). In 2012, masonry had its peak market share for newly registered construction projects in the UK, making up about 72% of the market (NHBC Foundation, 2016) and its similar figure in many developing and developed countries globally. Despite the extensive use of masonry in historical and contemporary constructions worldwide, previous image-based crack detection research has predominantly focused on infrastructure elements such as roads and bridges, primarily concentrating on concrete and asphalt surfaces, with comparatively less attention given to cracks in masonry (Munawar et al., 2021). This lack of research can be partially attributed to the shortage of publicly available datasets concerning masonry materials (Ali et al., 2022; Huang et al., 2024). Therefore, there remains a pressing need to establish datasets for masonry crack detection. Meanwhile, advanced algorithms are also needed to accomplish detection tasks.

Recent advancements in computer vision models offer promising avenues. The Segment Anything Model (SAM) was introduced by Meta AI (Kirillov et al., 2023). It is pre-trained on the largest-ever dataset (SA-1B) generated by a data engine, comprising 11 million images and over one billion masks, making it 400 times larger than the existing OpenImage dataset (Benenson et al., 2019). SAM has the ability to segment any object in any image after receiving prompts without requiring additional training. This highlights its exceptional capacity for generalization when handling diverse images and objects. The aim of SAM is to develop a powerful foundation model for segmentation, designed as a prompt model without active recognition ability. The model is currently undergoing extensive testing across various domains such as medical imaging, remote sensing, agriculture, 3D point cloud segmentation, and more, yielding remarkable results following specific modifications (Zhang et al., 2023). Therefore, this recent breakthrough is highly valuable for conducting experiments in the field of Structural Health Monitoring (SHM), by using this foundational visual model and developing further enhancements for the specific application. In this paper, our aim is to develop image-based automated models for identification and localization of bricks and cracks in masonry. The models are built on SAM as a state-of-the-art (SOTA) computer vision model, which we enhance by two novel methods for automation and trained using a self-built dataset.

To achieve this goal, the initial step involves enabling SAM for autonomous operation, which encompasses restructuring SAM's architecture and automating prompt generation. The second step is enhancing the quality of mask generation for detecting bricks and cracks in masonry by fine-tuning the original large-scale model. This process integrates LoRA, a technique within the SOTA Parameter-Efficient Fine-Tuning (PEFT) methodology (Hu et al., 2021). PEFT is used to fine-tune the SAM's encoder, and the decoder is also unfrozen for fine-tuning. Utilizing SAM as a foundation and incorporating transfer learning with the generalization abilities of pre-trained models, we carry out customized training by using a specialized, self-built instance segmentation dataset. This approach aims to yield advanced models proficient in detecting cracks specifically in masonry materials. Simultaneously, we employ SOTA instance segmentation algorithms to evaluate the proposed dataset, establishing benchmarks, and comparing them against the two SAM-based algorithms proposed in this paper.

Furthermore, previous research has emphasized the need for future crack studies to focus more on crack classification and measurement (Ali et al., 2022; Munawar et al., 2021). Hence, in this paper, we evaluated our model on a case study of a masonry wall.

Building on successful brick and crack detection, we propose a strategy based on Hough Line Transform and perspective transformation to estimate the actual crack dimensions automatically using known brick unit dimensions. The results are validated using a laser scans. To sum up, by establishing the dataset, developing SOTA SAM-based algorithms, and proposing a method for estimating masonry crack sizes, we have elevated the level of visual-based automated damage detection in masonry material, as well as the capability to quantify the severity of damage. The self-built masonry crack dataset, named MCrack1300, will be made publicly available at (upload later).

## 2. Literature review

### 2.1 Crack detection algorithms

Image-based crack detection methods have been extensively studied for various civil engineering applications (Mohan and Poobal, 2018; Munawar et al., 2021). They are essential non-destructive inspection techniques that analyse images of a structure's surface to extract and segment crack regions. They can be categorized into image processing methods and machine learning methods (Munawar et al., 2021). Image processing methods use filters (Fujita and Hamamoto, 2011; Lins and Givigi, 2016; Zalama et al., 2014), morphological analysis (Merazi-Meksen et al., 2014; Zhang et al., 2014), statistical techniques (Lins and Givigi, 2016), and percolation-based approaches (Wang and Huang, 2010; Yamaguchi and Hashimoto, 2010) for crack detection without involving a model training phase (Mohan and Poobal, 2018). Most of these algorithms rely on manual feature extraction, which, while more intricate and lightweight, is highly sensitive to factors like image quality, noise, and shadows (Mohan and Poobal, 2018; Munawar et al., 2021). Machine learning methods applied for crack detection can be categorized into traditional machine learning, such as Support Vector Machines (SVMs) (Chen et al., 2022), logistic regression (Yoo and Kim, 2016), random forest (Shi et al., 2016), boosting (Yang et al., 2020), clustering (Huyan et al., 2020; Lei et al., 2018), and deep learning, including Convolutional Neural Networks (CNN) like GoogleNet (Rao et al., 2021), VGG (Ali et al., 2021; Rao et al., 2021), U-Net (Liu and Wang, 2022; Zhang et al., 2021), ResNet (Fan et al., 2022), Mask R-CNN (Huang et al., 2022), etc.

Recently, networks based on the Vision Transformer (ViT) architecture have gained popularity, often offering performance enhancements. Some researchers have started integrating this technology into crack detection as well. Shamsabadi et al. (2022) achieved an increase of at least 3.8% in mean Intersection over Union (mIoU) using TransUNet, which is based on a CNN-ViT backbone, across multiple datasets compared to models using CNN-based architectures like DeepLabv3+ and U-Net. They discovered that models based on the ViT architecture demonstrated greater robustness against noise signals. Quan et al. (2023) conducted various comparative experiments on four road crack semantic segmentation datasets, where Transformer-based methods outperformed CNN-based ones on most test sets.

For image-based crack detection in masonry, there is considerably less research available. Dais et al. (2021) combined U-Net and Feature Pyramid Network (FPN), integrating various backbones such as VGG, ResNet, and MobileNet, to conduct comparative experiments specifically focused on pixel-level crack segmentation for masonry structures. They provide a dataset comprising 240 images (224×224) of cracks in masonry material for semantic segmentation and demonstrated the effectiveness of CNN-based methods. Huang et al. (2024) developed the Crack900 dataset, a collection specifically designed for crack detection in masonry, aligning RGB images with thermal imaging data. This dataset was subjected to tests with 11 different models, encompassing both CNN-based and ViT-based architectures. The study established benchmarks of the dataset and highlighted that employing a cascading strategy significantly enhances segmentation accuracy. Loverdos and Sarhosis (2022) explored CNN-based models for detecting cracks on masonry, and also studied crack shapes and generation of geometric models of masonry structures. They introduced engineering-focused ways to measure crack area and errors, showcasing the CNN methods' strengths over thresholding in image processing techniques. As a limitation, they caution that datasets for deep learning need irregular masonry for better crack detection, and for accurate geometric assessments, orthorectified images are essential. Minh Dang (2022) utilized some CNN-based algorithms for crack detection and proposed a method to determine brick unit sizes using the Mask R-CNN instance segmentation algorithm. Then, they perform a perspective transformation based on the detected four corner points of the complete known-sized brick unit to calculate the real dimensions of each pixel, and to extract actual crack length measurements. However, they find that this basic perspective transformation method may lead to inaccurate crack length measurements when applied to highly skewed input images (Minh Dang et al., 2022).

Overall, researchers continue to employ and enhance new algorithms to improve crack detection accuracy. Compared to image processing methods, machine learning methods, particularly deep learning, have gained popularity in recent years due to their higher accuracy and stronger resistance to interference. However, they heavily rely on high-quality datasets. Meanwhile, the lack of sufficient datasets remains a challenge, especially for masonry structures. Furthermore, integrating additional engineering applications such as dimension assessment, digital environmental geometric modelling, and others holds significant research potential.

## 2.2 Segment Anything Model

The architecture of SAM is shown in Figure 1. It includes three parts:

1. Image encoder part, which is used for extracting feature information from the input images, constructed from Masked AutoEncoders (MAE) (He et al., 2022) pre-trained ViT.
2. Prompt encoder part, which can accept two types of prompts: sparse (points, bounding box and text) and dense (masks).
3. Lightweight mask part, which can efficiently map the image embeddings, prompt embeddings, and an output token to a mask. The original version outputs a binary mask along with its corresponding bounding box.

Both prompt encoder and mask decoder are relatively light compared to the huge image encoder. Only ~50 ms (i.e., almost in real-time) is required in a web browser environment to run both the prompt encoder and mask decoder on a standard CPU.

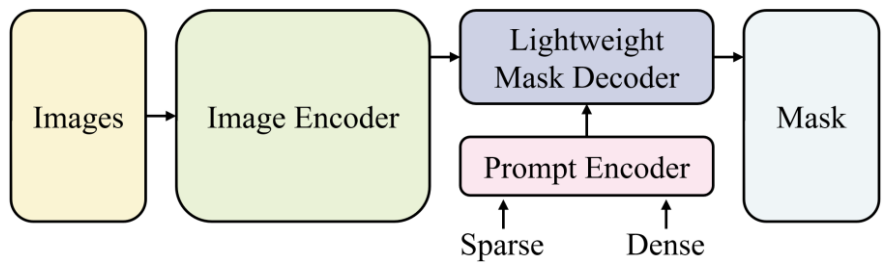

Figure 1. Basic architecture of the Segment Anything Model (Kirillov et al., 2023).

Semantic segmentation can theoretically be infinitely subdivided. This is shown in Figure 2, where a prompt point is placed in a laser image, at the top-left corner of the column, and then multiple valid masks are generated based on this prompt point. For the given example, SAM generates three masks, as the model perceives that this solution will encompass the majority of cases, namely, "whole", "part", and "subpart". This implies that using singular prompts may lead to ambiguity. Using additional prompt points and negative/background prompt points can resolve such ambiguities, obtaining the mask relevant to the user's needs. Therefore, the automated generation of prompts holds great research potential. Entering an appropriate prompt set can very likely directly obtain the required mask. Additionally, the zero-shot capability of SAM is not only prominently exhibited in RGB images but also extends to other image modalities, such as infrared (Chen and Bai, 2023), laser imagery (Figure 2), etc.

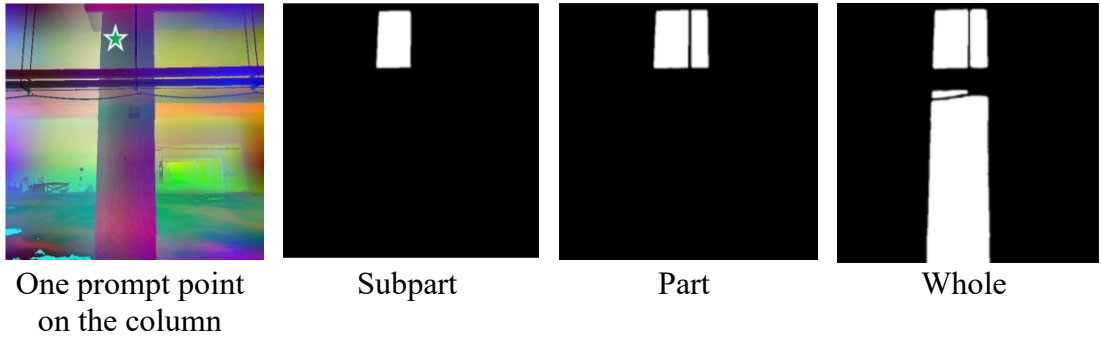

One prompt point on the column | Subpart | Part | Whole

Figure 2. An example involving a laser image with one prompt point and the corresponding valid masks.

### 2.3 Enhanced SAM-based models

#### 2.3.1 SAM-based methods for automatic mask generation

Several SAM-based methods have been proposed for automatic mask generation, and they can be roughly divided into the three mainstream methods shown in Figure 3.

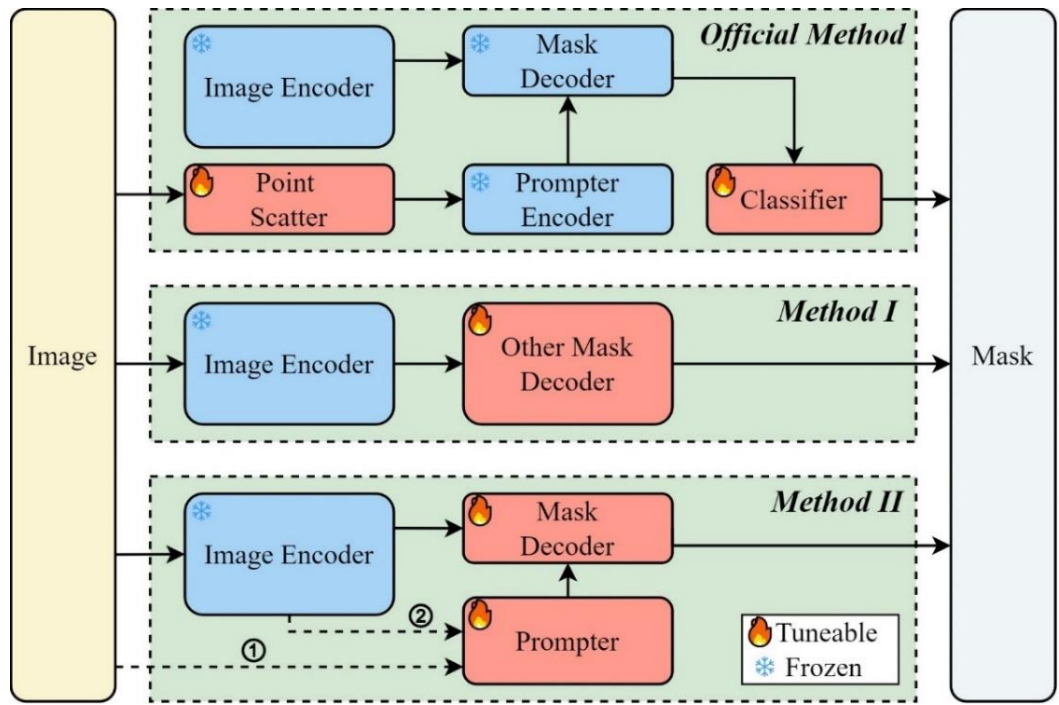

Figure 3. Three mainstream SAM-based methods for automatic mask generation.

As shown in Figure 3, the snowflake and flame symbols represent that model parameters in this section are frozen and tuneable, respectively. The official method is the "everything mode" of SAM shown in Figure 4, where the model evenly scatters points ($n \times m$, where $n$ and $m$ are user-customizable) on the image, generates numerous masks based on each prompt point, and then feeds the masks' corresponding targets into the classifier, which is trained by Contrastive Language-Image Pre-training (CLIP) (Radford et al., 2021). The output includes masks corresponding to objects with confidence scores surpassing the designated threshold (Kirillov et al., 2023). One drawback of this automated method is that the prompt points are uniformly distributed, which might not always be optimal. Additionally, due to the strong zero-shot capabilities of CLIP, SAM's "everything mode" generally performs well in segmenting common objects. However, it might not be suitable for the requirements of certain specialized downstream tasks like crack detection (Ahmadi et al., 2023).

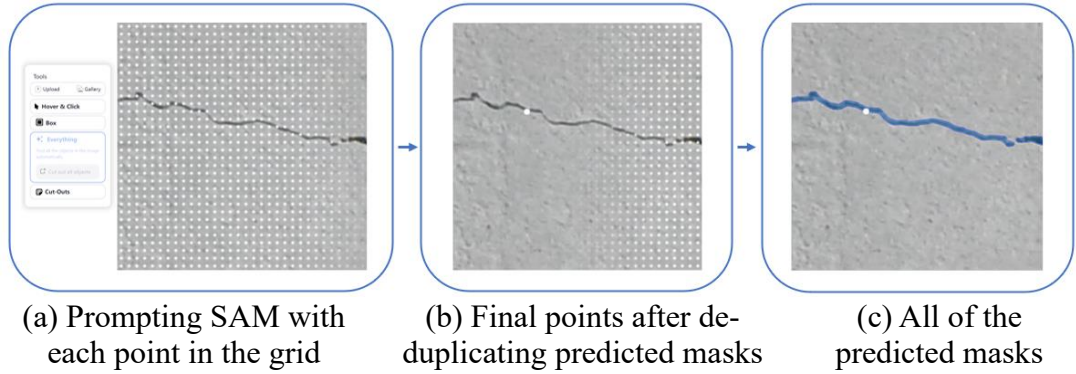

(a) Prompting SAM with each point in the grid  (b) Final points after de-duplicating predicted masks  (c) All of the predicted masks

Figure 4. Everything mode of SAM from official website ("Meta AI," 2023).

Clearly, the official method involves freezing the entire SAM and achieving automatic mask generation by externally adding additional components. Figure 3 illustrates two additional methods (*Method I* and *Method II*) for automated execution that are based on enabling SAM to adapt to downstream tasks, while disabling the training of the image encoder. *Method I* involves redesigning the mask decoder to function independently from the prompt encoder. *Method II* retains the original mask decoder and transforms the prompt encoder into a self-generating prompter to achieve automation.

*Method I* is a rather direct improvement that involves coupling the encoder of SAM with some other decoders, abandoning the need for prompt inputs, and achieving automated execution. This could entail directly linking the decoder of other algorithms or replacing the feature extraction (encoder) section of other instance segmentation algorithms like Mask R-CNN (He et al., 2018), Mask2Former (Cheng et al., 2022), etc. This hybrid approach enables the new model to benefit from SAM's robust encoder for automated tasks. Alternatively, it could use a different decoder or alter the task head to perform semantic segmentation (Ge et al., 2023) or panoptic segmentation (Nguyen et al., 2023). For instance, CrackSAM (Ge et al., 2023) fine-tunes the encoder while removing the prompt encoder and replacing the prompt embedding with learnable token, modifying the original mask decoder to automatically execute semantic segmentation tasks for cracks. Ultimately, when tested for zero-shot capabilities across various crack segmentation datasets, it achieved Intersection over Union (IoU) improvements ranging from 2.1% to 11.1% compared to the current best-performing SegFormer (Xie et al., 2021) model.

*Method II* is focused on modifications of the prompt encoder to enable the self-generating prompting function. This modified section is termed the "prompter" in this paper, and the two dashed lines illustrate two different prompter training approaches. Additionally, as the mask decoder is relatively lightweight, it is typically unfrozen and involved in the training process. Specifically, it can be achieved by introducing an additional object detection algorithm, such as Faster R-CNN (Kirillov et al., 2023; Ren et al., 2016), or some lightweight or rough sematic segmentation algorithm on specified objects, like pre-trained CNNs (Mayladan et al., 2023). Such independent algorithms can obtain the detected bounding boxes, prompt points or masks trained based on input images (dashed line 1 in *Method II* as shown in Figure 3), which serve as inputs to SAM's prompt encoder, thus acquiring the desired masks.

Another instance involves a self-prompt unit, which is trained by using image embeddings obtained from the SAM encoder (dashed line 2 in *Method II* as shown in Figure 3), along with the resized ground truth label. This unit predicts an initial mask, which is subsequently employed to determine the bounding box and location point for prompting SAM (Wu et al., 2023). This approach eliminates the need for introducing an additional neural network and directly uses the existing image encoder to extract features. Furthermore, researchers proposed abandoning the step of acquiring prompt boxes/points before feeding them into the prompt encoder. RSPrompter (K. Chen et al., 2023) was designed to capture features from the intermediate layers of the image encoder for feature fusion, and then use two types of prompter, anchor-based and query-based, to generate prompt embeddings for the SAM's mask decoder. This direct training method on prompt embeddings, compared to training on prompt points or boxes, minimizes the loss of high-dimensional information interaction. This approach, applied in remote sensing, showed an average 3% improvement in mean Average Precision (mAP) over the current SOTA algorithms across three different remote sensing datasets for segmentation performance evaluation (K. Chen et al., 2023).

Overall, as the current largest visual foundational model, SAM rapidly gained widespread attention after its release, leading to extensive research. Two SAM-based automated execution methods incorporate various alterations, varying in depth, to the mask decoder and prompt encoder, respectively, each demonstrating significant potential.

**2.3.2 Fine-tuning of SAM**

Given the widespread application of the large-scale foundational model, the concept of PEFT has garnered significant attention. This is primarily because fully training large models is a resource-intensive endeavour. Additionally, fine-tuning typically leads to some performance improvements for specific tasks. There are two mainstream directions for fine-tuning large models. The first approach involves selecting a small subset of the original model parameters for fine-tuning (Guo et al., 2021). For example, in SAM, the mask decoder component is relatively lightweight with only 4.7 MB, so most improvements will involve fine-tuning this particular part.

The second approach is to add some layers into the original model, and freeze the whole model parameters, only training parameters of additional layers. One typical approach is to add lightweight adapters (Houlsby et al., 2019) to the original model and adjust only the parameters of these adapter layers during fine-tuning (T. Chen et al., 2023; Ge et al., 2023; Pan et al., 2022; Wang et al., 2023). Adapters come in various forms, classified as either series (Figure 5(a)) or parallel (Figure 5(b)) structure, or a combination of both simultaneously. They can be flexibly inserted at different layers. The adapter layer (Figure 5(c)) is a small bottleneck structure, consisting of two feedforward neural networks (the former for dimensionality reduction and the latter for dimension restoration) and a non-linear layer within them.

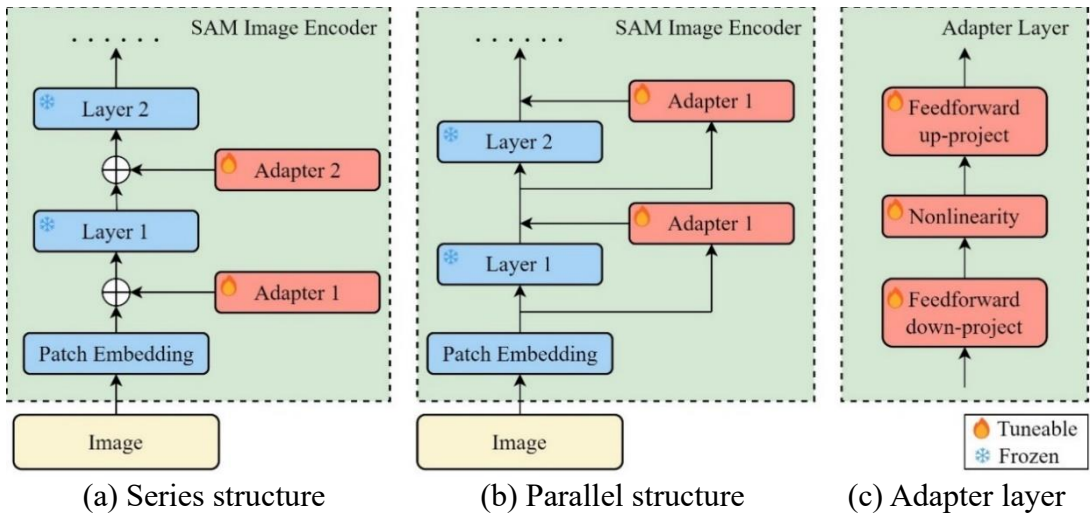

(a) Series structure     (b) Parallel structure     (c) Adapter layer

Figure 5. Diagram of adapter fine-tuning structure.

One example is SAM-Adapter (T. Chen et al., 2023), in which two Multilayer Perceptrons (MLPs) and an activation function within these MLPs are employed to construct the adapter layers, where the parameters of the first MLP is unshared and the second is shared. These adapter layers are attached to each transformer layer of the SAM's image encoder. The performance reached the SOTA level across multiple datasets. Another example is CrackSAM (Ge et al., 2023), which employs both parallel and series structures. Adapter layers are inserted sequentially after the attention layers and simultaneously inserted in parallel at the MLP. The Gaussian Error Linear Unit (GELU) activation function serves as the non-linear layer. This resulted in an improvement of IoU of crack from around 0.55 to approximately 0.65.

In addition to conventional adapter fine-tuning methods, an entirely different fine-tuning approach, called LoRA has been proposed recently, utilizing low-rank matrices to update parameters (Hu et al., 2021). Aghajanyan et al. (2020) suggest that the learned over-parameterized models exist on a lower intrinsic dimension. Therefore, LoRA freezes pre-trained weights and indirectly trains dense layers in neural network during task adaptation by optimizing their corresponding rank decomposition matrices. In this way, LoRA's parallel structure does not impact inference efficiency, and the linear fine-tuning matrix can be easily integrated into the original projection matrix, requiring no changes in network structure.

SAMed (Zhang and Liu, 2023) adopts this approach, implementing the LoRA module on each transformer block of the SAM image encoder, aiming to offer a universal solution for medical image segmentation. CrackSAM (Ge et al., 2023) incorporates LoRA to fine-tune SAM's encoder, more precisely the query and key weight matrices in the attention layer, for performing semantic segmentation tasks on cracks. This enhancement results in an IoU increase from 0.55 to 0.647 on the test set.

Besides, there are some other methods, such as HQ-SAM (Ke et al., 2023), which injects a learnable High-Quality Output Token (less than 0.5% of the parameter of whole SAM) into SAM's mask decoder. $44 \times 10^3$ high quality masks were used to train the token to enhance SAM for high-quality zero-shot segmentation. Overall, due to the extensive parameter space of large-scale models, comprehensive fine-tuning is unfeasible, necessitating the utilization of PEFT techniques. This technology holds large potential, enabling targeted domain adaptation for specific downstream tasks by harnessing the superior prior knowledge of large-scale models with limited resources, leading to significant improvements.

The conventional adapter introduces additional layers, resulting in inference latency (Hu et al., 2021), and for attaining similar effectiveness, the parameter count of the adapter layer is typically larger than LoRA (Hu et al., 2023). Moreover, methods introducing extra learnable tokens rely on more extensive and richer dataset guidance and training. Thus, considering the application scenario of this paper, choosing LoRA as the fine-tuning technique is more appropriate.

**2.4 Instance segmentation**

The original SAM performs the process from prompt to mask and then calculates the corresponding bounding boxes based on the boundaries of mask. Additionally, its segmentation possesses the ability to distinguish between different instances. Therefore, performing instance segmentation tasks based on SAM allows for better retention of information originating from SAM. Figure 6 depicts the ground truth for various computer vision tasks for a given image. Figure 6(b) illustrates SAM's everything mode, demonstrating its understanding of the overall scene, such as the table separated by wires (pink), as well as its ability to comprehend different instances, such as mice.

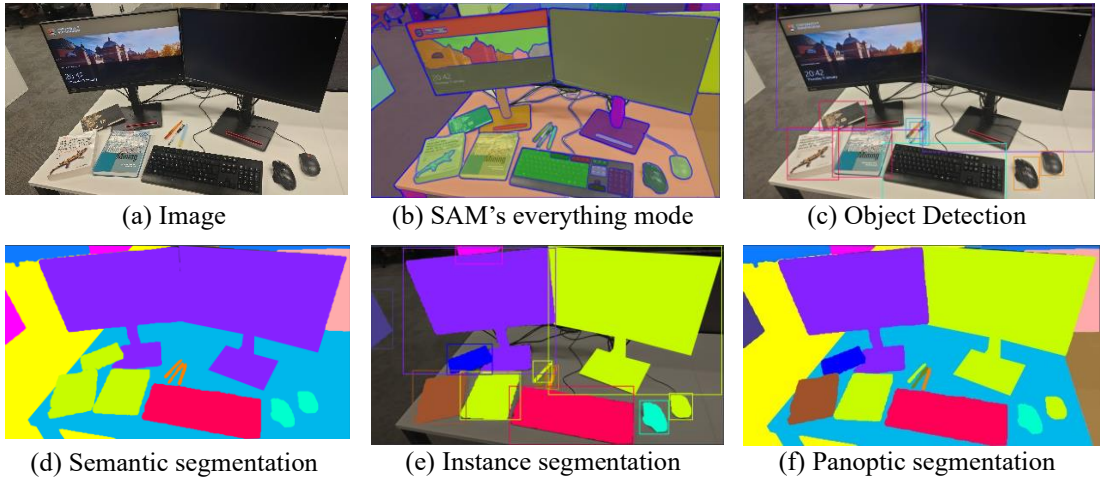

(a) Image         (b) SAM's everything mode         (c) Object Detection

(d) Semantic segmentation         (e) Instance segmentation         (f) Panoptic segmentation

Figure 6. Ground truth exhibited by different computer vision tasks for image (a)

Currently, approaches for instance segmentation can be broadly categorized into two groups: top-down methods, which are detection-based, and bottom-up methods, which are segmentation-based (Chen et al., 2019a; Gu et al., 2022).

The detection-based method is similar to an object detection task. In this method, the initial step is to identify the region for each individual instance and subsequently segment the instance mask within that specific area. Since Mask R-CNN (He et al., 2018) was published, which developed from the two-stage Faster R-CNN (Ren et al., 2016) object detector by adding a parallel mask prediction branch, it has been the dominant method for instance segmentation. Other representative algorithms in Mask R-CNN series include Mask Scoring (MS) R-CNN (Huang et al., 2019), which integrates a mask IoU branch into the Mask R-CNN framework to evaluate the quality of segmentation; Cascade Mask R-CNN (Cai and Vasconcelos, 2021), which introduces a cascading architecture with multiple stages to gradually refine and improve segmentation accuracy; and Hybrid Task Cascade (HTC) (Chen et al., 2019a), which also employs a cascaded architecture and interweaves detection and segmentation tasks for a joint multi-stage processing, while introducing a global information guiding mechanism.

The second method is segmentation-based one that predicts the category label for each pixel and then employs a clustering or bounding detection technique to group these pixels together, thereby generating the results for instance segmentation. In comparison to the former method, this approach is more intricate. Representative algorithms include YOLACT (Bolya et al., 2019), which is achieved by generating prototype masks and predicting coefficients, and then linearly combining to produce the final instance masks. Segmenting Objects by Locations (SOLO) treats instance segmentation as a dense prediction problem by directly performing segmentation at each pixel position, and achieves predictive accuracy comparable to Mask R-CNN (Wang et al., 2020a). SOLOv2 adopts the concept of dynamic convolution, leading the process of object mask generation decoupled into a mask kernel prediction and mask feature learning (Wang et al., 2020b).

Building on the extensive use of Transformers (Vaswani et al., 2017), Detection Transformer (DETR) (Carion et al., 2020) was proposed as the first object detection framework to implement Transformers in downstream computer vision tasks. DETR revolutionized conventional object detection methodologies by entirely discarding anchor-based and non-maximum suppression modules, and instead introduced a set prediction approach. In this approach, each object is treated as a query to generate paired masks and labels for the objects. MaskFormer (Cheng et al., 2021) adopts the modelling and training methodologies from DETR, framing segmentation tasks as mask classification challenges, but with slower convergence. Based on that, Mask2Former (Cheng et al., 2022) incorporates masked attention to confine cross-attention to foreground regions, notably accelerating network training speed. As a universal image segmentation, Mask2Former achieves SOTA performance across tasks including semantic segmentation, instance segmentation, and panoptic segmentation. It can be seen that such query-based unified segmentation frameworks have great prospects. SAM also adopts such a design, where the input prompt embedding to the mask decoder is concatenated with a token, which serves the same purpose. Therefore, subsequent improvements will also revolve around such methods.

Based on the discussion above, research on masonry crack detection is limited, and there is a lack of datasets. Additionally, existing image-based methods for estimating masonry crack sizes are inadequate for handling complex conditions. The promptable SOTA visual foundation model SAM has tremendous potential to achieve optimal detection, while it requires substantial modifications to enable automated detection tasks. The solutions to these gaps will be discussed in the next section.

## 3. Methodology

In this section, we initially introduce the masonry crack dataset we developed. Subsequently, we present two innovative, enhanced SAM-based automated instance segmentation algorithms, following *Methods I* and *Method II* as summarized in the literature review. This encompasses further exploration of the original SAM, fine-tuning of the SAM encoder using LoRA, establishment of the feature extractor, and detailed architectures description of two enhanced methods, *Method I\**: LoRA-SAM with Mask2Former decoder and *Method II\**: LoRA-SAM with self-generating prompter. Here, we use "\*" to denote specific implementation methods following a particular improvement. After that, we introduce the instance segmentation evaluation method utilized in this paper. Finally, we demonstrate the masonry crack size evaluation method based on the Hough Line Transform and perspective transformation.

### 3.1 Dataset creation

For this paper, we created a new dataset for instance segmentation containing 1300 images of masonry cracks, each with a resolution of 640 pixels × 640 pixels. We call the dataset "MCrack1300". Out of these, 1000 images were allocated for the training set, 150 for the validation set, and 150 for the testing set. There are a total of five annotated labels: bricks, broken bricks, cracks, spalling, and plant. The quantities of the latter two categories are relatively low and were not included in the actual training. According to previous research (Loverdos and Sarhosis, 2022), a distinction was made between regular bricks and irregular bricks caused by breakage or obstruction (labelled as "broken brick"), enabling deep neural networks to autonomously differentiate between them, facilitating their use in subsequent engineering application tasks. The data sources are as follows: 251 images from Crack900 as RGB images (Huang et al., 2024), 376 images from the internet, and 673 images collected using a mobile phone from the University of Birmingham UK campus and its vicinity. The dataset contains various types of bricks with different colours, textures, and materials, along with cracks of varying sizes and shapes, including small, longitudinal, transverse, and step-like forms. The dataset was annotated using the online labelling tool provided by the Roboflow (2020), allowing for the visualization of annotated results online. The annotation method is a closed polyline encompassing the mask. Users have the flexibility to choose the annotation file format for download, such as COCO, VOC, YOLO, and more, suiting their specific needs. Figure 7 displays some examples from

MCrack1300.

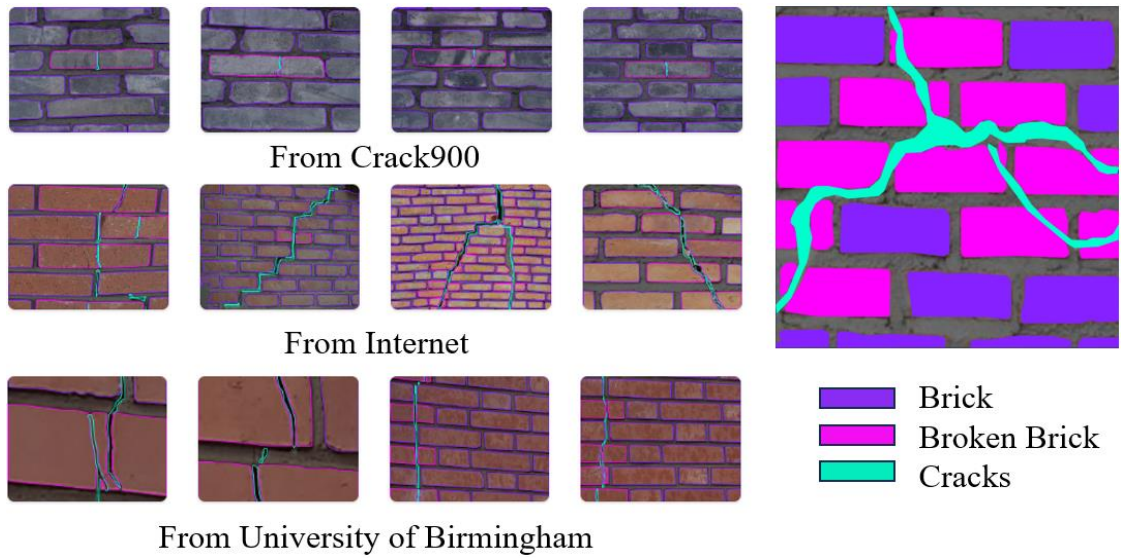

Figure 7. Some examples from our MCrack1300 dataset.

For the training set, the distribution of the number of instances per image is shown in Figure 8. The mode, median, and mean of the number of instances for each image are 11, 13 and 15.87. The maximum number of instances per image does not exceed 60.

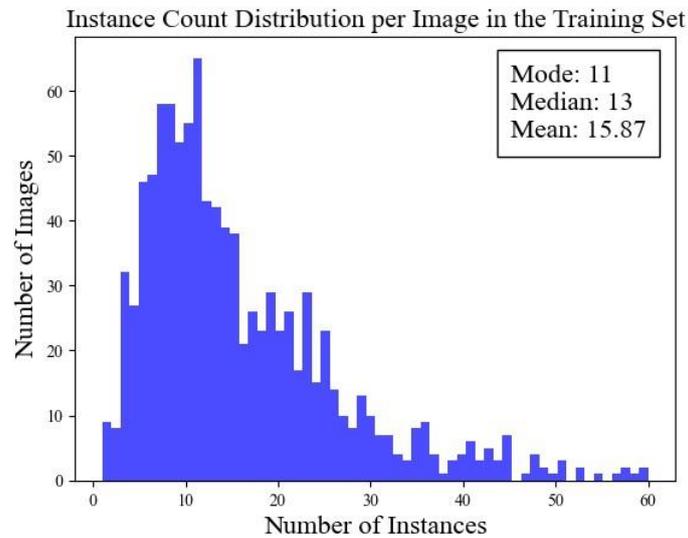

Figure 8. Distribution of the training set.

**3.2 Enhanced SAM-based automated instance segmentation algorithms**

**3.2.1 Fine-tuning SAM's encoder using LoRA**

The detailed SAM architecture is displayed in Figure 9.

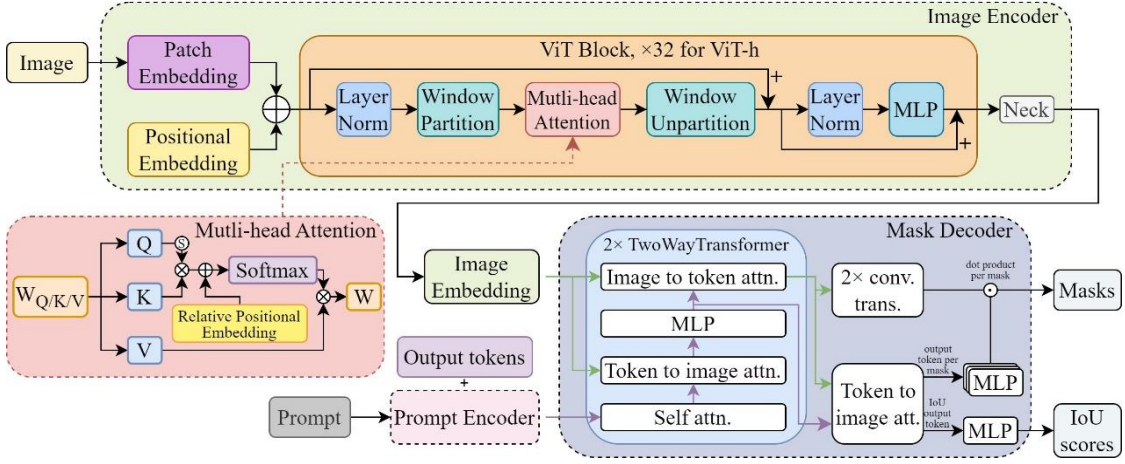

Figure 9. Detailed SAM architecture reproduced from (Kirillov et al., 2023)

As shown in the original SAM architecture (Figure 9), the image encoder is an MAE (He et al., 2022) pre-trained ViT, consisting of a patch embedding layer, a learnable positional embedding, a ViT backbone, and a refined neck. The most computationally intensive part is the multi-head attention layer within the ViT block. The multi-head attention mechanism of SAM are expanded, and the computation of scaled dot-product attention process (Vaswani et al., 2017) is adjusted based on MViTv2 (Li et al., 2022b) as follows:

$$Attention(Q, K, V) = softmax\left(\frac{QK^T}{\sqrt{d_k}} + pos_{rel}\right)V \quad\quad Equation\ 1$$

where *Q, K,* and *V* represent the three vectors in the *Attention* mechanism, namely Query, Key, and Value, used to compute attention distributions. Softmax means activation function. $K^T$ represents the transpose matrix of *K*. $d_k$ is the dimension of input queries and keys, which aids in managing the scope of attention scores. Based on MviTv2, it also incorporates decomposed relative position embedding ($pos_{rel}$). The backbone consists of stacked ViT blocks. To ensure optimal performance, in this paper, we only use ViT-Huge (ViT-H), with an embedding dimension, number of blocks, and number of attention heads set to 1280, 32, and 16, respectively. The neck is responsible for reducing the dimension to 256 for image embedding.

In our proposed enhancements of SAM, first we implement fine-tuning on the extensive encoder of SAM using the LoRA method. Specifically, this fine-tuning involves the Q, K, and V matrices within the attention layer as discussed above. Figure 10 shows the schematic diagram of LoRA's structure.

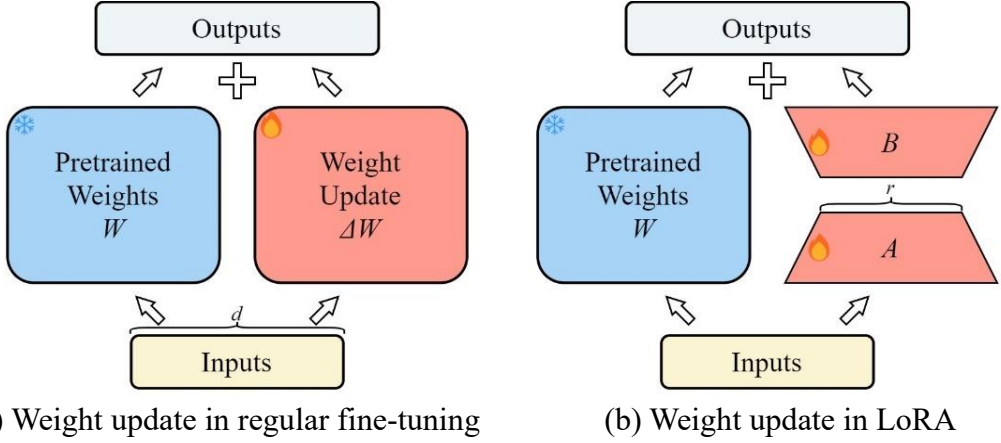

(a) Weight update in regular fine-tuning  (b) Weight update in LoRA

Figure 10. Comparison between regular and LoRA fine-tuning (Hu et al., 2021).

In Figure 10, *A* and *B* are LoRA matrices, which represent the weight update matrix *ΔW*; *r* is the inner dimension/rank of LoRA matrix, which is a hyperparameter typically set to values such as 4, 8, or 16 etc. << *d* (the dimension/rank of the input vector); and the size of the pre-trained weight matrix *W* is *d×d*. The typical weight updating process proceeds as follows:

$$W_{updated} = W + \Delta W \qquad \text{Equation 2}$$

As shown above, the decomposition of *ΔW* implies representing the large matrix *ΔW* with two smaller LoRA matrices, A and B. We can express the decomposition as *ΔW = AB* (where *AB* denotes the matrix product of *A* and *B*). In this way, the number of parameters significantly decreases from the original *d×d* to *d×r+r×d*. Therefore, from a matrix perspective, LoRA can be applied to any weight matrix within a neural network, enabling effective fine-tuning with a small number of parameters.

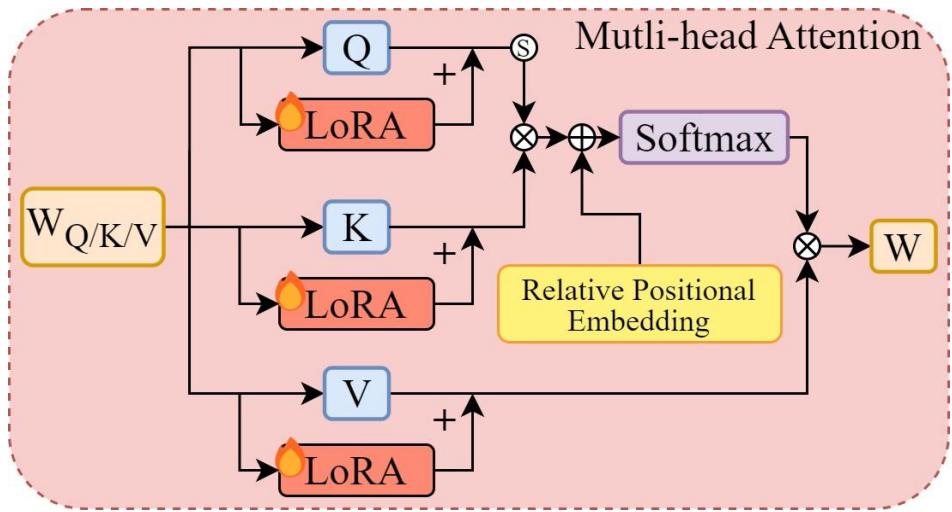

Figure 11. LoRA fine-tuning on Q, K, and V matrices of multi-head attention layers

In this paper, shown in Figure 11, to facilitate a more comprehensive fine-tuning of the SAM's encoder, we focus on the crucial and computationally intensive Q, K, and V matrices within the multi-head attention layers. This allows the SAM's encoder to adapt to the attention distribution for our tasks. The fine-tuning process is applied to all 33 multi-head attention layers within the ViT backbone, utilizing LoRA-decomposed matrices.

**3.2.2 Feature extractor**

The backbone of SAM is a non-hierarchical ViT that consistently maintains a single-scale feature map. However, to address multi-scale object detection in a downstream task, we still need to construct multi-scale feature maps. These serve as inputs for the decoder in *Method I* and as inputs for training the prompter in *Method II*. Li's (2022a) research on object detection indicates that in a non-hierarchical ViT, the final layer concentrates the most information. Constructing multi-scale feature maps from this last layer outperforms the method of using intermediary layers imitating FPN for constructing multi-scale feature maps. Therefore, we propose a five-stage multi-scale simple feature pyramid structure derived directly from the final layer of the ViT backbone.

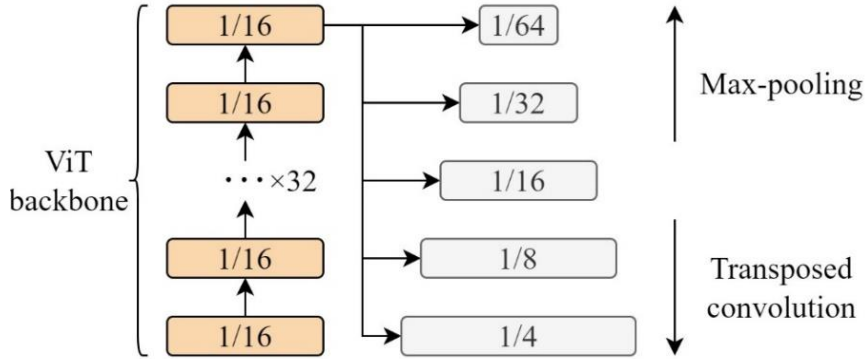

Figure 12. Five-stage multi-scale simple feature pyramid structure.

As shown in Figure 12, the final ViT backbone layer's feature map undergoes learnable transposed convolutions for up-sampling and max-pooling for down-sampling to create a simple feature pyramid. This notation *1/n* (*n* = 4, 8, 16, 32, 64) refers to the feature map size being *W/n* and *H/n* relative to the original input image dimensions *W* and *H*. To obtain a 1/4-sized feature map, two transposed convolutions with a kernel size of 2 and stride of 2 are used successively, separated by a normalization layer and GELU activation. Down-sampling to a 1/64-sized feature map is achieved through max-pooling operations from a 1/32-sized feature map. The output channels are adjusted to 256 to match the subsequent step.

For the decoder requiring input in *Method I*, each of the obtained five-scale feature maps undergoes a sequential 1×1 convolution, normalization layer, and a 3×3 convolution with a stride of 1, reducing channels to 256 before inputting to the decoder. For the feature map to be fed into *Method II's* prompter, we initially compressed the channels to 256, following the acquisition of the final layer feature map from the ViT. Then, the feature pyramid building process is executed, still maintaining channels at 256 when inputting to the prompter. Because we believe the training for prompter might demand fewer features, we initially reduce the channels to conserve computational resources.

### 3.2.3 Method I*: LoRA-SAM with Mask2Former decoder

Since the inspiration behind SAM's mask decoder draws from DETR (Carion et al., 2020) and MaskFormer (Cheng et al., 2021), executing per-mask classification and modifying the standard Transformer decoder. It naturally suggests combining the decoder of Mask2Former (Cheng et al., 2022) with SAM's image encoder to achieve the automatic execution of SAM. This method follows the *Method I* as discussed in Section 2.3.1, and the proposed architecture is shown in Figure 13.

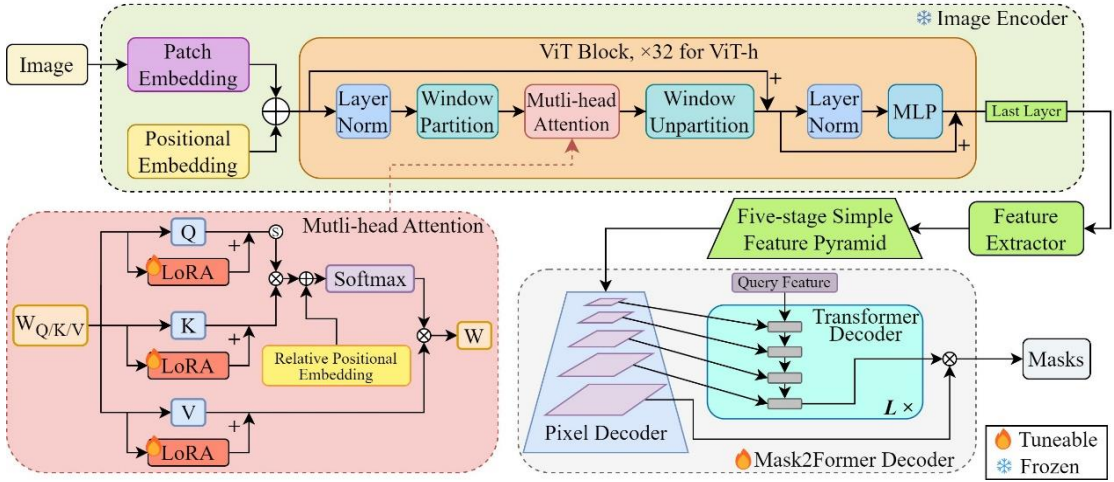

Figure 13. Architecture of LoRA-SAM with Mask2Former Decoder (*Method I**).

The prompt and original mask decoder are replaced by the Mask2Former decoder (Cheng et al., 2022). Image features from SAM's encoder are fed to the Mask2Former decoder to complete instance segmentation tasks automatically. The heavy encoder of SAM is fine-tuned using LoRA technology, as described in Section 3.2.1. This process involves fine-tuning the query, key, and value matrices within the multi-head attention layers, shown in Figure 13. The original features fed into the Mask2Former decoder are derived from the final layer of the ViT backbone, which pass through the feature extractor, and a five-stage simple feature pyramid is constructed as mentioned in Section 3.2.2.

Inside the Mask2Former decoder (Cheng et al., 2022), there is a pixel decoder alongside a Transformer decoder, utilizing the Deformable Transformer (Zhu et al., 2021) for multi-scale feature interaction. Each stage of the multi-scale feature pyramid is sequentially fed into one layer of the Transformer decoder and repeated *L* times, as shown in Figure 13. Alternating between low and high-resolution features in the Transformer decoder allows the network to better capture objects at different scales without introducing excessive computational overhead.

Specifically, in the pixel decoder, apart from the largest-sized layer (*W/4, H/4*), the remaining multi-scale feature maps undergo channel unification before being fed into the Transformer decoder (as we have done in feature extractor). Then, for each pixel on these *n* feature maps, *k* points are predicted by forecasting *k (x, y)* offsets using the current pixel's position as the reference point at each resolution. A five-stage pyramid is used in our paper *(n=5)*, one stage more than the original version because it can enhance the recognition ability of crack, which is well-suited for our data (see Section 4.2.3). Therefore, utilizing the features of *(n-1)k* points, the current pixel's features are updated to achieve cross-feature-map localized information interaction, and reduce the computational load from *(H×W)*$^2$ to *HW×k* compared to computing attention across global pixel features. For the largest-sized feature map, feature fusion occurs with up-sampled features from the 1/8 resolution feature map, directly utilized for high-resolution feature prediction in mask inference.

For the Transformer decoder, it sequentially comprises mask attention, self-attention layers, and a Feed-Forward Network (FFN) (Cheng et al., 2022). Its input includes the image features discussed before, as well as learnable query features and masks. The mask attention confines network attention within each query's corresponding mask, enabling the model to focus on specific local Regions of Interest (ROI). Its ultimate output are query embeddings, which, along with the high-resolution feature maps that underwent feature fusion as discussed before, perform cross-product operations to predict masks. The loss computation method remains consistent with Mask2Former (Cheng et al., 2022) during the training process, including the mask loss calculation, which also involves computing with few randomly sampled points.

### 3.2.4 Method II*: LoRA-SAM with self-generating prompter

This section will discuss how to follow *Method II* (see Section 2.3.1) to establish an automatically executable SAM. Clearly, as discussed in the literature review section, introducing an additional detection network solely to extract information from images and then incorporating detection results into the prompt is redundant and cumbersome. Hence, the feature extractor described in Section 3.2.2 is employed here to directly extract features from SAM's robust encoder for training a self-prompting mechanism (K. Chen et al., 2023). Additionally, based on the previous discussion of the original SAM architecture, training to generate the prompt token for the SAM's mask decoder is the most direct method, which can minimize loss in high-dimensional information interaction compared to training aimed at prompt points/box/mask.

SAM's mask decoder is retained. The input of the mask decoder includes image embeddings and the prompt token. The detailed mask decoder is depicted in Figure 9, and for further details, refer to (Kirillov et al., 2023). Overall, the cross-attention mechanism in the mask decoder allows for the interaction between image embedding and token, resulting in the generation of masks. Here, we select the highest-scoring mask among the generated multiple masks. The proposed architecture is shown in Figure 14.

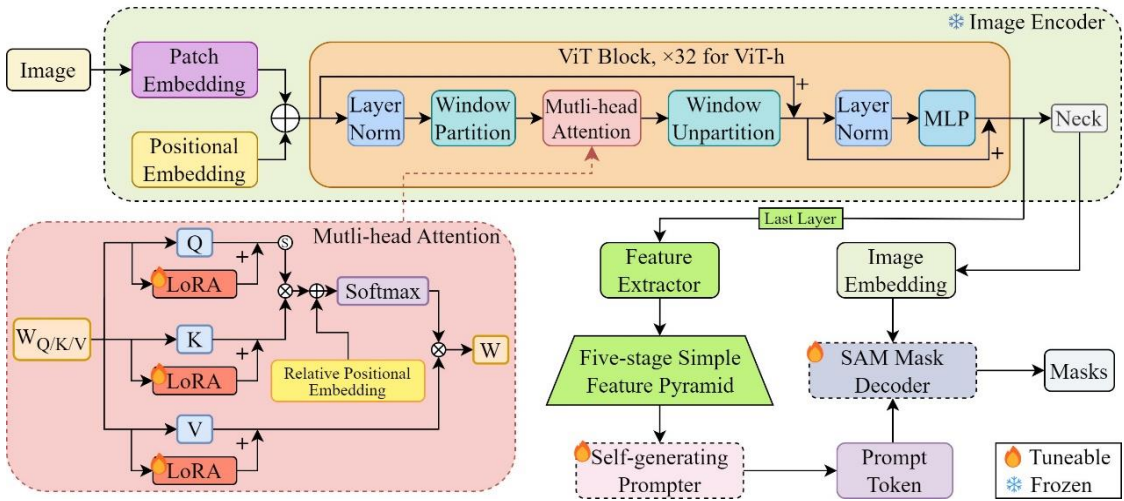

Figure 14. Architecture of LoRA-SAM with self-generating prompter (*Method II*\*).

As we can see in Figure 14, the components involved in the training include the LoRA module within the multi-head attention layer of the ViT backbone in the SAM encoder, running in parallel to the query, key, value matrices, the self-generating prompter, and the mask decoder of the SAM. The entire heavy SAM's encoder is frozen during the training process.

We use a query-based prompter, primarily composed of a Transformer encoder and decoder (K. Chen et al., 2023), whose output meets the input requirements of the SAM's mask decoder. The five-stage simple feature pyramid from the feature extractor mentioned in Section 3.2.2 is fed into the Transformer encoder layer to perform self-attention. The decoder transforms predefined learnable queries into prompt embeddings for SAM, achieved by cross-attention with the results from the encoder and learnable tokens initialized as zero. Two hyperparameters, *Np* and *Kp*, need to be introduced. Here, *Np* represents the number of prompt groups, indicating the count of instances, while *Kp* defines the number of embeddings per prompt. Therefore, *Np* corresponds to the number of prompt tokens for SAM's mask decoder.

The loss function during training comprises two parts. The first step is matching, which involves aligning N predicted masks with M ground truth masks, where N is greater than M. This step employs the Hungarian matching algorithm (Kuhn, 2010), consistent with algorithms like DETR (Carion et al., 2020) and Mask2Former (Cheng et al., 2022). Once each predicted instance has been matched with its corresponding ground truth, supervised information can be obtained. This mainly involves multi-class classification and binary mask classification, as described below:

$$Loss = \frac{1}{N_p}\sum_{i}^{N_p}(L_{cls}^i + \alpha L_{seg}^i) \qquad \text{Equation 3}$$

where, $L_{cls}^i$ indicates the cross-entropy loss computed between the predicted category and the target. $L_{seg}^i$ indicates the binary cross-entropy loss between the predicted mask and the corresponding ground-truth instance mask after matching. $\alpha$ is an indicator to confirm a positive match.

### 3.3 Evaluation method

The evaluation method for instance segmentation models uniformly adopts the widely acknowledged COCO (Microsoft Common Objects in Context) official evaluation criterion for instance segmentation tasks (Lin et al., 2014). A positive prediction is classified as true when the IoU between the predicted instance's box/mask and its respective ground truth surpasses a threshold *T*, and the predicted category aligns with the ground truth category. In this paper, our primary focus lies on six evaluation metrics: $AP_{box}$, $AP_{box}^{50}$, $AP_{box}^{75}$, $AP_{mask}$, $AP_{mask}^{50}$, $AP_{mask}^{75}$. The Average Precision (*AP*) is computed across various IoU values, typically 10 IoU thresholds ranging from 0.50 to 0.95 with increments of .05. This diverges from the conventional method where *AP* is calculated at a single IoU of .50 (referred to as metric $AP^{50}$). Averaging across IoU values benefits detectors that exhibit superior localization. Moreover, AP is computed as an average across all categories, commonly referred to as "mean average precision" (*mAP*). *AP* uses the concept of recall and precision to calculate (Lin et al., 2014):

$$Recall = \frac{TP}{TP+FN} \qquad \text{Equation 4}$$

$$Precision = \frac{TP}{TP+FP} \qquad \text{Equation 5}$$

$$AP = \frac{1}{101}\sum_{r\in\{0,0.01,\ldots 1\}} p_{interpolation}(r) \qquad \text{Equation 6}$$

Where *TP* means true positives; *FN* means false negatives; *FP* means false positives; $p_{interpolation}(r)$ is the precision obtained through interpolation at the given recall level *r*, and precision is averaged in the set of 101 equally spaced recall levels [0.0, 0.01, 0.02 …. 1.0].

$AP_{mask}$ will be considered the most crucial metric in this paper as we prioritize segmentation capability. A higher *AP* signifies more accurate predicted instance box/mask, indicating superior instance localization/segmentation performance. $AP^{50}$ denotes evaluation under the IoU threshold of 0.50, while $AP^{75}$ is a stricter metric evaluated under an IoU threshold of 0.75. Therefore, $AP^{75}$ outperforms $AP^{50}$ in evaluating box/mask accuracy.

### 3.4 Size estimation of masonry cracks

In previous studies (Minh Dang et al., 2022), the method to estimate crack size involved detecting the four corners of bricks for perspective transformation to calculate the real dimensions of each pixel. However, the four corners of detected masks for bricks are often not sufficiently clear, while the four corners of detected bounding boxes are used as substitutes (Minh Dang et al., 2022), leading to inaccuracies in perspective transformation calculations. Therefore, they acknowledge that their method is limited to accommodating only a finite range of skewed shooting angles. We argue that utilizing the common edge lines of multiple bricks placed side by side or in parallel is a better approach, due to more detected edge pixels involved in this process. Therefore, taking advantage of the simplicity in identifying horizontal and vertical perspective lines in brick images, following from our proposed instance segmentation algorithms for bricks and cracks, we introduce a monocular-based automated method to obtain the perspective transformation matrix of an image. It can transform images and predicted masks into orthographic projection maps, allowing for the accurate estimation of detailed crack dimensions, given the known real sizes of brick units. The core of accurate perspective transformation relies on obtaining four points with high precision from the original image. These four points should correspond to a rectangle on the orthographic image, and the aspect ratio of this rectangle needs to be known. Based on this, the perspective transformation matrix can be calculated accurately. The general workflow is illustrated in Figure 15.

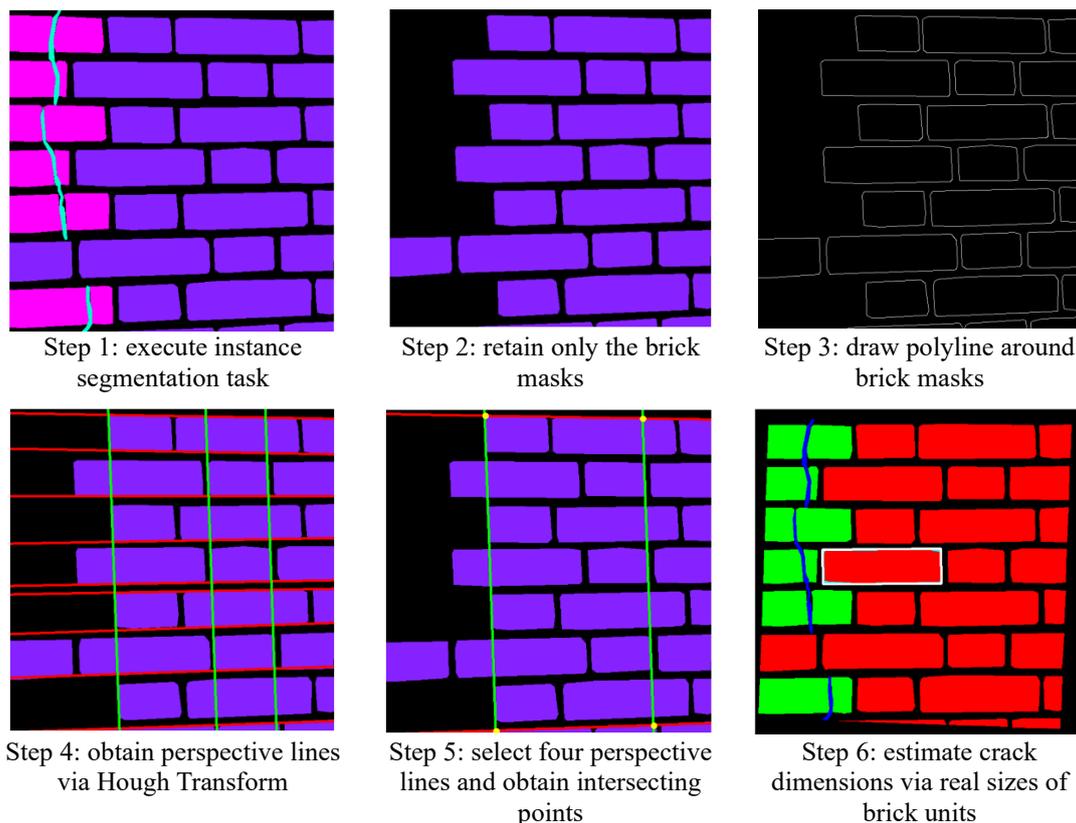

Figure 15. Workflow of estimating crack size based on known real sizes of brick units.

Initially, the instance segmentation task is executed to identify instances of cracks, bricks, and broken bricks (Step 1). Subsequently, all masks of cracks and broken bricks are removed, leaving only the brick mask. This facilitates better identification of perspective lines using the Hough Line Transform (Illingworth and Kittler, 1988), avoiding interference from other masks. The outcome after removal is shown in Figure 15 (Step 2). The Hough Transform (Illingworth and Kittler, 1988) maps image pixels into polar coordinate, representing lines with the polar equation, shown below:

$$r = x\cos(\theta) + y\sin(\theta) \qquad \qquad Equation\ 7$$

Therefore, we need to simplify the image, retaining only the edge pixels of the bricks. Since the predicted masks are represented by enclosing polygons, polygons around these masks can be drawn directly as shown in Step 3 in Figure 15. For other formats of masks, a format conversion can be performed, or edge detection be performed. Then, for every pixel in the image of Step 3, we traverse the parameter space (i.e., $\theta$ and $r$ in polar coordinates), computing the corresponding $r$ and $\theta$ values, and accumulating in the Hough space. Such high-accumulated points represent potential lines in the image.

In Step 4, we set two different thresholds for the horizontal and vertical perspective lines (using 45° as a divider). The threshold for detecting horizontal lines should typically be higher than that for vertical lines, nearly by factor two, due to the usual horizontal arrangement of bricks that maintains a consistent pattern every other layer. This implies that the range of high-accumulated points for horizontal direction is generally larger than that for vertical pixel accumulation values. Finally, the selected accumulated points ($r$ and $\theta$) transform from Hough space back into the image space to obtain the detected horizontal (red) and vertical (green) perspective lines' positions.

Step 5 is to select two pairs of perspective lines that are farthest from each other in vertical and horizontal directions. This approach helps to minimize errors as much as possible. Considering that the extracted lines on the plane generally have very similar slopes for each direction, this can be simplified to calculate the difference in intercepts. Then, four intersecting points are obtained based on four such lines. Afterwards, the perspective transformation can be executed based on these four points, converting the image into an orthographic projection. In the transformed orthographic image, these four points form the corners of a rectangular shape. The ratio between the vertical and horizontal sides of the transformed rectangle is set to be equal to the ratio between the left and bottom sides of the quadrilateral formed by the four points on the original, untransformed image as an initial value, awaiting correction later.

Upon executing the perspective transformation and obtaining the orthographic image,

the pixel-to-real-world scale in the horizontal and vertical directions will be inconsistent. To address this, there is the need to extract the mask of the bricks once more, calculate the aspect ratio of the bricks, shown in Step 6, and based on the known brick unit size ratio, restore the image's original aspect ratio. Ultimately, obtaining the real-world dimensions corresponding to each pixel in the orthographic image allows for the calculation of crack dimensions and area.

## 4. Implementation and testing

In this section, we apply the two previously proposed methods (*Method I**: LoRA-SAM + Mask2Former decoder and *Method II**: LoRA-SAM with self-generating prompter) on our MCrack1300 dataset. Our methods are compared with SOTA instance segmentation algorithms to establish benchmarks. Subsequently, we conduct ablation experiments to investigate how different factors influence our methods. Lastly, we present a case study that involves bot instance segmentation and the automatic recognition of brick crack sizes.

### 4.1 Model Implementation

#### 4.1.1 Implementation detail

The implementation details are shown in Table 1.

Table 1. implementation details during SAM-based model training.

| Parameter | Setting | Explanation |
|---|---|---|
| Query number | 80 | For *Method I**, it should be suitable for the instance count per image, which is less than 60 shown in Figure 8. For error redundancy and computational load reduction, the value was set slightly higher than the maximum number of instances in a single image. |
| ($N_p$, $K_p$) | (80,8) | For *Method II**, $N_p$ represents the number of prompt groups, while $K_p$ signifies the number of embeddings per prompt. The setting of the $N_p$ value and the choice of query number are motivated by similar reasons. |
| Input size | 1024×1024 | Maintained from the original design of SAM |
| Optimizer | AdamW | From (Loshchilov and Hutter, 2017a) |
| Learning strategy | Cosine Annealing | A linear warm-up strategy for initial 50 iterations, followed by setting the learning rate to 1e-4, gradually decreasing to 1e-7 (Loshchilov and Hutter, 2017b) |
| Epoch | 400 | Obtained through experiments |
| Mini-batch size | As large as possible | For *Method I** is 16 (4 per GPU), and for *Method II** is 4 (1 per GPU) |
| Backbone | ViT-H | We need to extract features from the most informative layer, so the deeper the network, the better the performance |
| Data augmentation | Large Scale Jitter (LSJ) | Following the same strategy with Mask2Former (Cheng et al., 2022), it includes randomly sampling scales within the range of 0.1 to 2.0, followed by fixed-size cropping to the original size (1024×1024) |

| Acceleration techniques | DeepSpeed ZeRO 2 | Using Zero Redundancy Optimizer (ZeRO) stage 2 with FP16 (Rajbhandari et al., 2020) for saving GPU memory and accelerating computations |

We compare the proposed methods against a series of SOTA instance segmentation algorithms, all sourced from OpenMMLab (Chen et al., 2019b). For each algorithm, we selected the best-performing model configurations within their respective series, based on pre-training results *mAP* on the COCO dataset (Lin et al., 2014). We use the default optimizer, learning strategy, and mini-batch size from each model's configuration file; and the epochs are increased to a suitable value for each algorithm. All models utilize initial weights pre-trained on the COCO dataset (Lin et al., 2014). We use the same data augmentation strategy as applied in the two proposed methods. All experiments were conducted using NVIDIA A100 Tensor Core GPUs. The software environment consists of MMCV 2.1 and PyTorch 2.0.1 with CUDA version 11.7.

### 4.1.2 Comparison with the SOTA

Table 2. Comparative Analysis of SOTA instance segmentation algorithms on our dataset using the COCO evaluator.

| Method | $AP_{box}$ | $AP_{box}^{50}$ | $AP_{box}^{75}$ | $AP_{mask}$ | $AP_{mask}^{50}$ | $AP_{mask}^{75}$ |
|---|---|---|---|---|---|---|
| Mask R-CNN (ResNeXt) (He et al., 2018) | 0.642 | 0.763 | 0.682 | 0.619 | 0.758 | 0.630 |
| Mask R-CNN (Swin) (He et al., 2018) | 0.643 | 0.774 | 0.688 | 0.612 | 0.759 | 0.623 |
| MS R-CNN (Huang et al., 2019) | 0.628 | 0.752 | 0.665 | 0.623 | 0.741 | 0.638 |
| HTC (Chen et al., 2019a) | 0.648 | 0.778 | 0.680 | 0.613 | 0.736 | 0.625 |
| YOLACT (Bolya et al., 2019) | 0.567 | 0.737 | 0.612 | 0.571 | 0.687 | 0.594 |
| SOLOv2 (Wang et al., 2020b) | 0.680 | 0.777 | 0.700 | 0.651 | 0.791 | 0.655 |
| Cascade Mask R-CNN (Cai and Vasconcelos, 2021) | 0.690 | 0.792 | 0.723 | 0.642 | 0.791 | 0.641 |
| Mask2Former (Cheng et al., 2022) | 0.691 | 0.795 | 0.723 | 0.667 | 0.822 | 0.668 |
| **LoRA-SAM + Mask2Former decoder** | **0.722** | **0.827** | **0.755** | **0.703** | **0.853** | **0.701** |
| **LoRA-SAM + self-generating prompter** | 0.696 | 0.799 | 0.713 | 0.695 | 0.851 | 0.686 |

The results from the best epoch are determined by $AP_{mask}$ on the validation set. SOLOv2 directly obtains instance segmentation results without calculating bounding boxes. The bounding boxes here are obtained based on the obtained masks and used for evaluation and comparison. It is evident that both our proposed methods have reached or exceeded the SOTA level, especially excelling in segmentation performance. Compared to the best benchmark from Mask2Former, *Method I\** achieves an improvement of 3.6%, while *Method II\** is 2.8% better.

Based on the criterion of optimal $AP_{mask}$, we selected three top-performing SOTA algorithms alongside two SAM-based methods proposed by use for comparison, conducting a thorough evaluation of specific metrics for each category. All models exhibited high and similar performance on relatively regular instances (bricks). However, on irregular instances, broken bricks and cracks, our SAM-based model demonstrated a notable advantage, especially concerning highly irregular instances like cracks. *Method I\** displayed significant advantages in localization and identification of cracks ($AP_{box}$), whereas both SAM-based methods exhibited better performance in crack segmentation compared to other algorithms.

Table 3. Comparative Analysis of SOTA instance segmentation algorithms on our dataset for specific categories.

| **Method** | Category | $AP_{box}$ | $AP_{box}^{50}$ | $AP_{box}^{75}$ | $AP_{mask}$ | $AP_{mask}^{50}$ | $AP_{mask}^{75}$ |
|---|---|---|---|---|---|---|---|
| SOLOv2 (Wang et al., 2020b) | Brick | 0.913 | 0.962 | 0.940 | 0.915 | 0.963 | 0.949 |
| | Broken Brick | 0.783 | 0.858 | 0.809 | 0.788 | 0.867 | 0.825 |
| | Crack | 0.343 | 0.511 | 0.352 | 0.250 | 0.544 | 0.191 |
| Cascade Mask R-CNN (Cai and Vasconcelos, 2021) | Brick | **0.920** | 0.962 | 0.951 | 0.917 | 0.963 | 0.950 |
| | Broken Brick | 0.793 | 0.876 | 0.841 | 0.797 | 0.822 | 0.844 |
| | Crack | 0.355 | 0.539 | 0.377 | 0.211 | 0.527 | 0.128 |
| Mask2Former (Cheng et al., 2022) | Brick | **0.920** | **0.976** | **0.955** | 0.925 | **0.978** | **0.958** |
| | Broken Brick | 0.790 | 0.865 | 0.827 | 0.803 | 0.881 | 0.843 |
| | Crack | 0.362 | 0.546 | 0.387 | 0.274 | 0.607 | 0.205 |
| LoRA-SAM + Mask2Former decoder | Brick | 0.919 | 0.959 | 0.944 | **0.934** | 0.971 | 0.957 |
| | Broken Brick | **0.828** | **0.899** | **0.854** | **0.841** | **0.908** | **0.877** |
| | Crack | **0.418** | **0.623** | **0.467** | 0.335 | 0.680 | **0.271** |
| LoRA-SAM + self-generating prompter | Brick | 0.917 | 0.956 | 0.934 | 0.931 | 0.968 | 0.949 |
| | Broken Brick | 0.773 | 0.839 | 0.789 | 0.820 | 0.883 | 0.850 |
| | Crack | 0.398 | 0.601 | 0.417 | **0.335** | **0.702** | 0.259 |

Some visualized inference results of comparative experiments are shown in Figure 16.

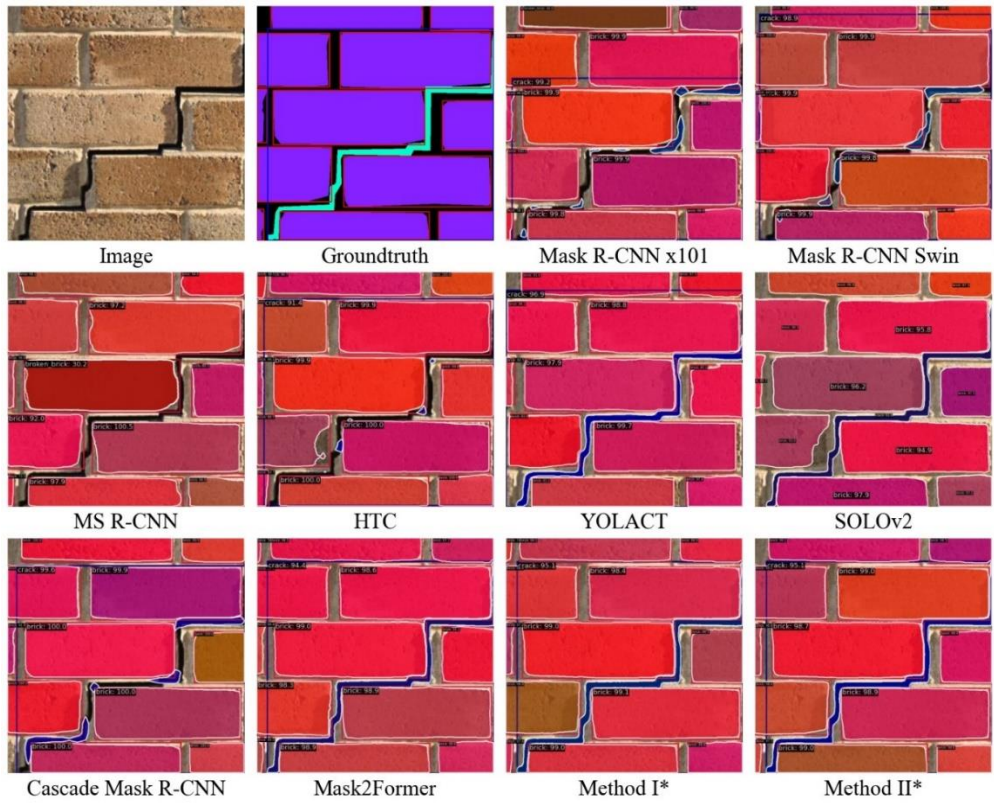

(a) Example of step crack in masonry.

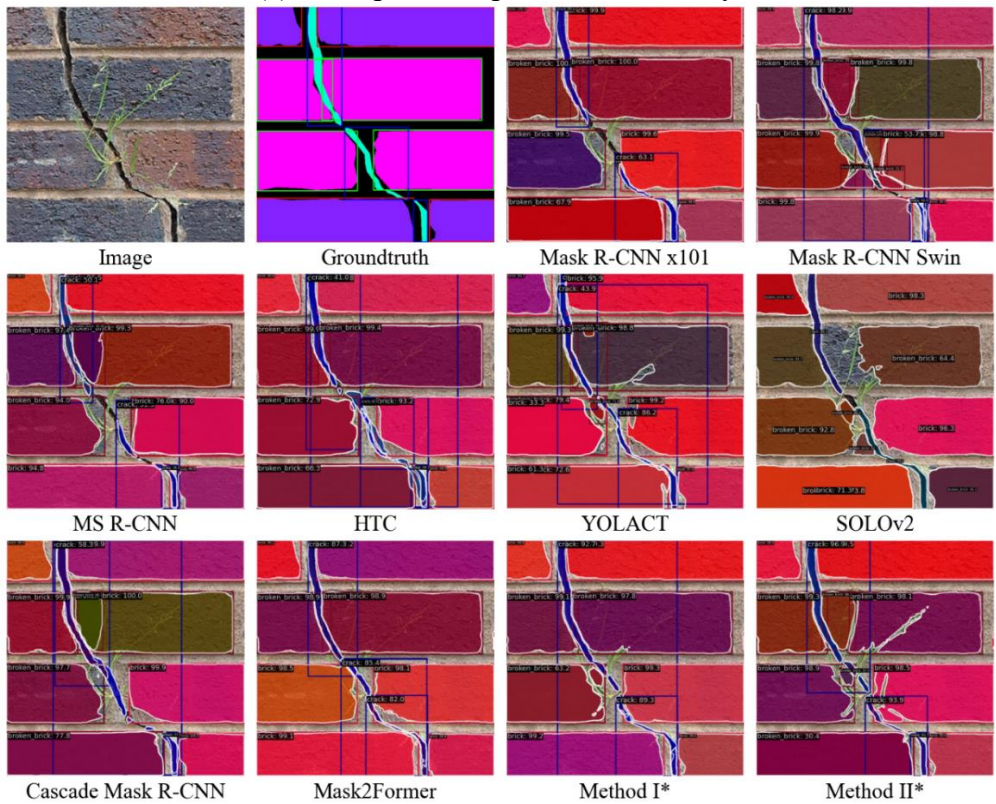

(b) Example of segmentation capability in the presence of interference.

Figure 16. Visualized inference results of comparative experiments.

The segmentation results for both horizontal and vertical cracks are generally satisfactory for most of the models. In the example, we present a relatively complex step crack, and the SAM-based methods still demonstrate stable recognition. Meanwhile, in scenarios with some interference, the advanced Cascade Mask R-CNN and Mask2Former have been affected. Observing the segmented edges of the two bricks near the plant in Figure 16(b), they significantly trail behind the SAM-based methods in detailed segmentation. In particular, the self-generating method exhibits remarkable detail segmentation capabilities.

**4.2 Ablation study**

or the collected dataset, *Method I\** slightly outperforms *Method II\**. Also, due to the transformer architecture introduced by the prompter, its training demands more resources. However, the potential of the self-generating prompter method remains undeniable. Moving forward, the focus will mainly be on conducting ablation experiments using *Method I\*:* LoRA-SAM with Mask2Former decoder as the example to investigate the influence of various factors on the SAM-based automated execution algorithm.

**4.2.1 Different ViT backbone**

There are three different sizes of ViT architectures: ViT-Base (ViT-B), ViT-Large (ViT-L) and ViT-Huge (ViT-H). The detailed parameters of different ViT architectures are displayed in Table 4.

Table 4. Parameters of different size of ViT backbone.

| **Backbone Size** | Embedding dimension | Blocks | Attention head | Params. |
|---|---|---|---|---|
| ViT-B | 768 | 12 | 12 | 83M |
| ViT-L | 1024 | 24 | 16 | 307M |
| ViT-H | 1280 | 32 | 16 | 632M |

Table 5 showcases the obtained results under different backbone sizes. As we only utilized the final layer as the original feature map, the impact of different backbone sizes on the results is quite significant. Deeper architectures yield better results.

Table 5. Comparative Analysis of different size of ViT backbone.

| Backbone Size | $AP_{box}$ | $AP_{box}^{50}$ | $AP_{box}^{75}$ | $AP_{mask}$ | $AP_{mask}^{50}$ | $AP_{mask}^{75}$ |
|---|---|---|---|---|---|---|
| ViT-B | 0.684 | 0.788 | 0.697 | 0.673 | 0.815 | 0.673 |
| ViT-L | 0.705 | 0.800 | 0.729 | 0.691 | 0.833 | 0.684 |
| ViT-H | 0.722 | 0.827 | 0.755 | 0.703 | 0.853 | 0.701 |

**4.2.2 The selection of number of queries**

Queries are input into the Mask2Former decoder. As mentioned in Section 3.2.4, the Hungarian matching algorithm (Kuhn, 2010) is adopted to align predicted masks with ground truth masks, and each query is matched to one mask. In the training set, each image has no more than 60 instances. Clearly, as shown in Table 6, a slight redundancy (queries=80) leads to better results. However, if the number of queries becomes too large, it not only increases GPU memory consumption and reduces computational speed, but also leads to worse outcome. As the number of queries increases, it is necessary to simultaneously increase total epochs to make sure the model converge to the optimal values, for example (queries = 100, epochs = 400) versus (queries = 100, epochs = 500). However, excessive epochs may deteriorate the results. The primary reason can be attributed to our use of the Cosine Annealing (Loshchilov and Hutter, 2017b) algorithm to adjust the learning rate. Too many epochs lead to a slow decrease in the learning rate, causing an unstable training process. On the other hand, too few epochs result in a rapid decline in learning rate, trapping the model in local optima. Therefore, the number of queries should well match with total epochs. A similar situation also occurred with *Method II\** when adjusting parameter $N_p$.

Table 6. Comparative analysis of different query numbers.

| Number of queries | $AP_{box}$ | $AP_{box}^{50}$ | $AP_{box}^{75}$ | $AP_{mask}$ | $AP_{mask}^{50}$ | $AP_{mask}^{75}$ |
|---|---|---|---|---|---|---|
| 60 | 0.704 | 0.813 | 0.723 | 0.697 | 0.848 | 0.692 |
| 60 (epochs = 500) | 0.704 | 0.811 | 0.724 | 0.690 | 0.837 | 0.685 |
| 80 | 0.722 | 0.827 | 0.755 | 0.703 | 0.853 | 0.701 |
| 100 | 0.703 | 0.809 | 0.723 | 0.692 | 0.841 | 0.687 |
| 100 (epochs = 500) | 0.715 | 0.819 | 0.728 | 0.698 | 0.853 | 0.689 |

### 4.2.3 Different feature extractor architectures

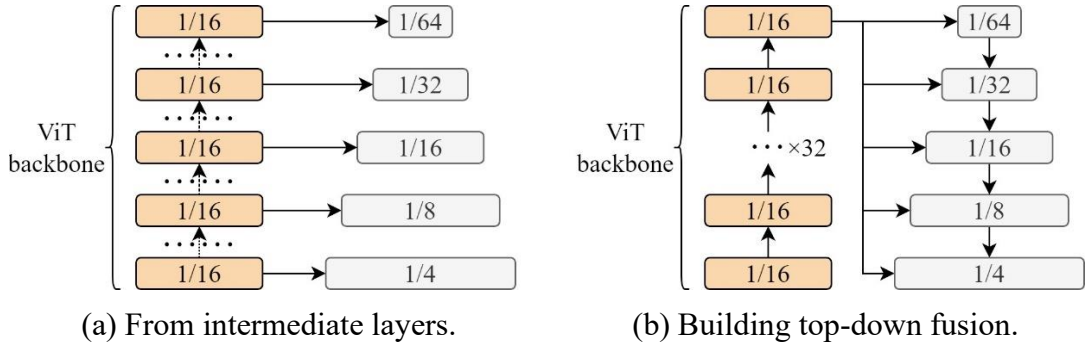

(a) From intermediate layers.  (b) Building top-down fusion.

Figure 17. Different feature extractor architectures.

We tested the impact of different feature extractors, including methods for constructing feature pyramids by extracting feature maps from the intermediate layers (7th, 13th, 19th, 25th, 32nd) of the ViT backbone (Figure 17(a)). The second method involved extracting features from the last layer of the ViT backbone to build the feature pyramid, and then from the top to the bottom, each layer is up-sampled and fused with the next layer by addition (Figure 17(b)). The third method employed a simple feature pyramid structure, and we tested two situations: four and five stages. The results are shown in Table 7. It shows that the effect obtained by constructing a feature pyramid using uniformly sampled intermediate layers from ViT backbone is not as good as using only the last layer, which is different from the feature pyramids built by traditional CNN networks but aligns with the conclusions of prior research (Li et al., 2022a) on non-hierarchical ViT backbone. After adding top-down fusion, the difference in performance compared to a directly constructed simple feature pyramid is not significant, but it increases the computation time by about 8%. Using the default four-stage feature pyramid mode of Mask2Former decoder can reduce memory consumption and decrease computation time by approximately 10%, but it results in a 1% decrease in performance.

Table 7. Comparative analysis of different feature extractor architectures.

| **Feature extractor** | $AP_{box}$ | $AP_{box}^{50}$ | $AP_{box}^{75}$ | $AP_{mask}$ | $AP_{mask}^{50}$ | $AP_{mask}^{75}$ |
|---|---|---|---|---|---|---|
| From intermediate layers | 0.694 | 0.794 | 0.721 | 0.684 | 0.828 | 0.680 |
| Building top-down fusion | 0.712 | 0.816 | 0.733 | 0.700 | 0.855 | 0.694 |
| Four-stage feature pyramid | 0.702 | 0.814 | 0.711 | 0.693 | 0.840 | 0.677 |
| Five-stage feature pyramid | 0.722 | 0.827 | 0.755 | 0.703 | 0.853 | 0.701 |

### 4.2.4 Impacts of LoRA

We tested the impact of employing LoRA to assess its utility, and the results indicate a significant benefit, approximately 4% increase in $AP_{mask}$. Also, we tested an important parameter, the rank value $r$, during LoRA fine-tuning, but this did not yield a noticeable change, especially in segmentation ability. The results are shown in Table 8.

Table 8. Comparative analysis of different LoRA parameters.

| Different LoRA param. | $AP_{box}$ | $AP_{box}^{50}$ | $AP_{box}^{75}$ | $AP_{mask}$ | $AP_{mask}^{50}$ | $AP_{mask}^{75}$ |
|---|---|---|---|---|---|---|
| No LoRA | 0.667 | 0.763 | 0.684 | 0.664 | 0.801 | 0.658 |
| LoRA (r = 8) | 0.711 | 0.811 | 0.735 | 0.698 | 0.843 | 0.691 |
| LoRA (r = 16) | 0.722 | 0.827 | 0.755 | 0.703 | 0.853 | 0.701 |
| LoRA (r = 32) | 0.715 | 0.820 | 0.738 | 0.700 | 0.853 | 0.701 |

### 4.2.5 Impacts of freezing the Mask Decoder

We used *Method II\** to test whether fine-tuning the SAM lightweight mask decoder could lead to improvements, and the results showed a notable enhancement, shown in Table 9.

Table 9. Comparative analysis of unfreezing SAM's mask decoder.

| SAM mask decoder | $AP_{box}$ | $AP_{box}^{50}$ | $AP_{box}^{75}$ | $AP_{mask}$ | $AP_{mask}^{50}$ | $AP_{mask}^{75}$ |
|---|---|---|---|---|---|---|
| Freezing | 0.630 | 0.736 | 0.652 | 0.613 | 0.743 | 0.622 |
| Unfreezing | 0.696 | 0.799 | 0.713 | 0.695 | 0.851 | 0.686 |

## 4.3 Case study

In this section, we conduct a case study. We used the Terrestrial Laser Scanner (TLS) to collect point cloud data along with RGB panoramic images from a masonry wall . Subsequently, we performed brick and crack instance segmentation based on RGB images, employing *Model I\**: LoRA-SAM with the Mask2Former decoder. After obtaining the results, we then executed the proposed automatic crack size recognition algorithm based on known brick unit dimensions.

### 4.3.1 Data collection

The data was collected at the University of Birmingham, and the dimensions of the entire wall are around 13.5 m width and 4.5 m height. The size of the brick unit is 220

×110×60 mm and 215×102.5×65 mm for vertical bricks.

We used a Faro Focus$^s$ Laser Scanner for collecting the point cloud, selecting the profile "Outdoor 20 m". The resolution was set to ½ and the quality was 2×. The straight-line distance to the wall during scanning was no more than 5 meters. We scanned three stations to enhance point cloud density and facilitate point cloud registration. Based on previous research (Oytun and Atasoy, 2022), the parameters we set provide sufficient redundancy to capture the cracks during scanning. The images used for performing the visual detection of cracks also come from panoramic images obtained from this scanner. The results of the scans and image are shown in Figure 17.

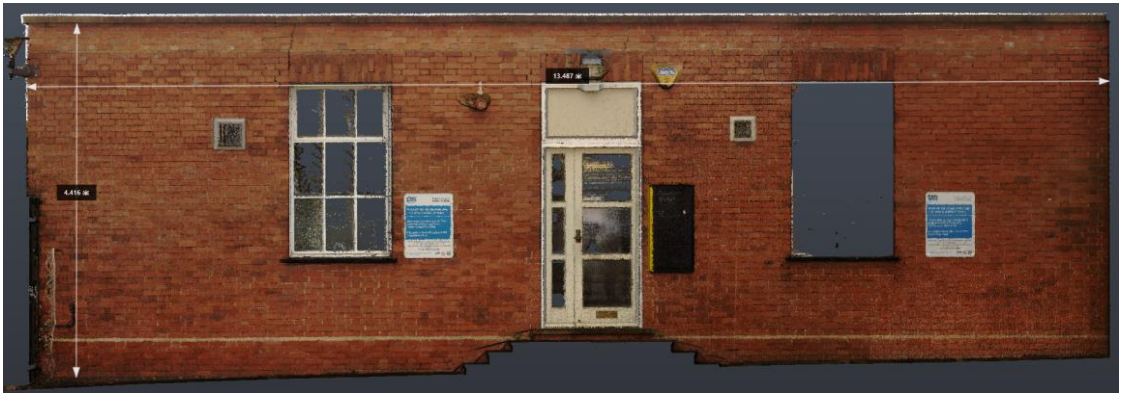

(a) Point cloud results of masonry wall.

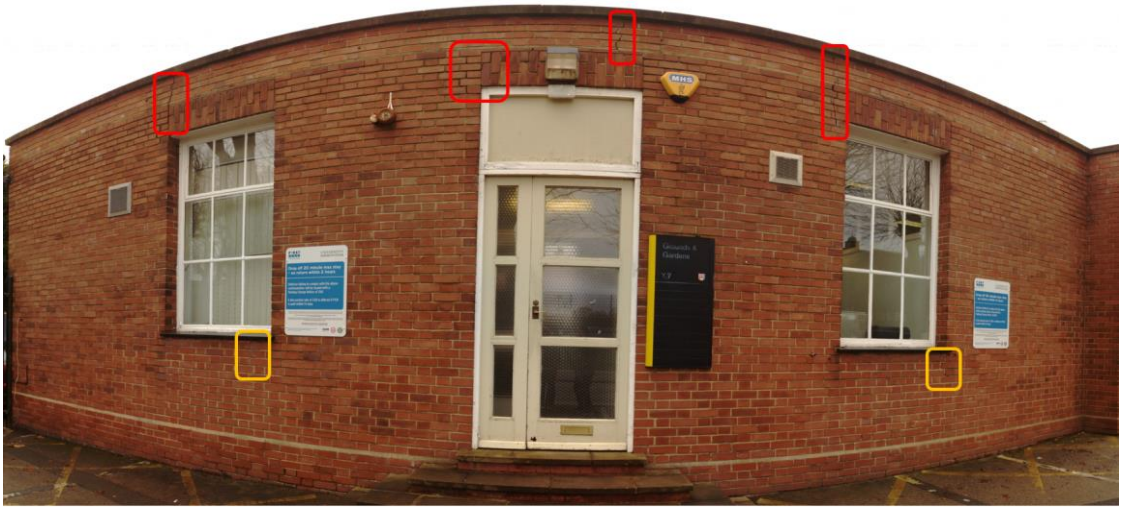

(b) Panoramic image of masonry wall.

Figure 18. Point cloud and panoramic image of the case study masonry wall.

Through visual inspection (see Figure 18(b)), we can clearly see that there are four noticeable cracks in the upper part of the wall. Additionally, in the lower portion, there are two less prominent, slight cracks near the windows on both sides.

**4.3.2 Model implementation for case study**

We initially employed the GroundSAM algorithm, which combines Grounding DINO (Liu et al., 2023) and SAM. This model is capable of detecting and segmenting anything with text inputs. However, it is limited to recognizing common objects. Using this algorithm, we identified doors, windows, and billboards on the brick wall and replaced them with white pixels (See Figure 19 (a)). Afterwards, we extracted sub-images from the panoramic image, averaging the cropping in 6 rows and 14 columns, as shown in Figure 19(b). In the end, we obtained a total of 84 sub-images with resolution is 497 × 524.

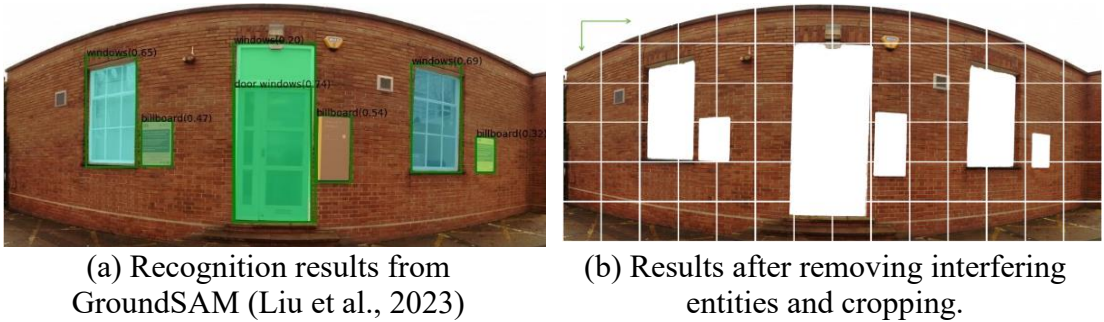

| (a) Recognition results from GroundSAM (Liu et al., 2023) | (b) Results after removing interfering entities and cropping. |

Figure 19. Data preprocessing for input to the model for case study.

Finally, we used *Method I\** and the weights learned from the data we have established for crack and brick detection, as well as the approach proposed in Section 3.4 to estimate crack dimensions.

**4.3.3 Result of crack dimension estimation**

For ease of viewing, instances of the same class are displayed in the same colour. The examples in the Figure 20 include the cropped input image to the model, the detection results, the perspective-transformed image, and approximate regions corresponding to the point cloud. We set the top-left corner of the cropped image as coordinates (1, 1), the first coordinate represents the row number, and the second one represents the column number.

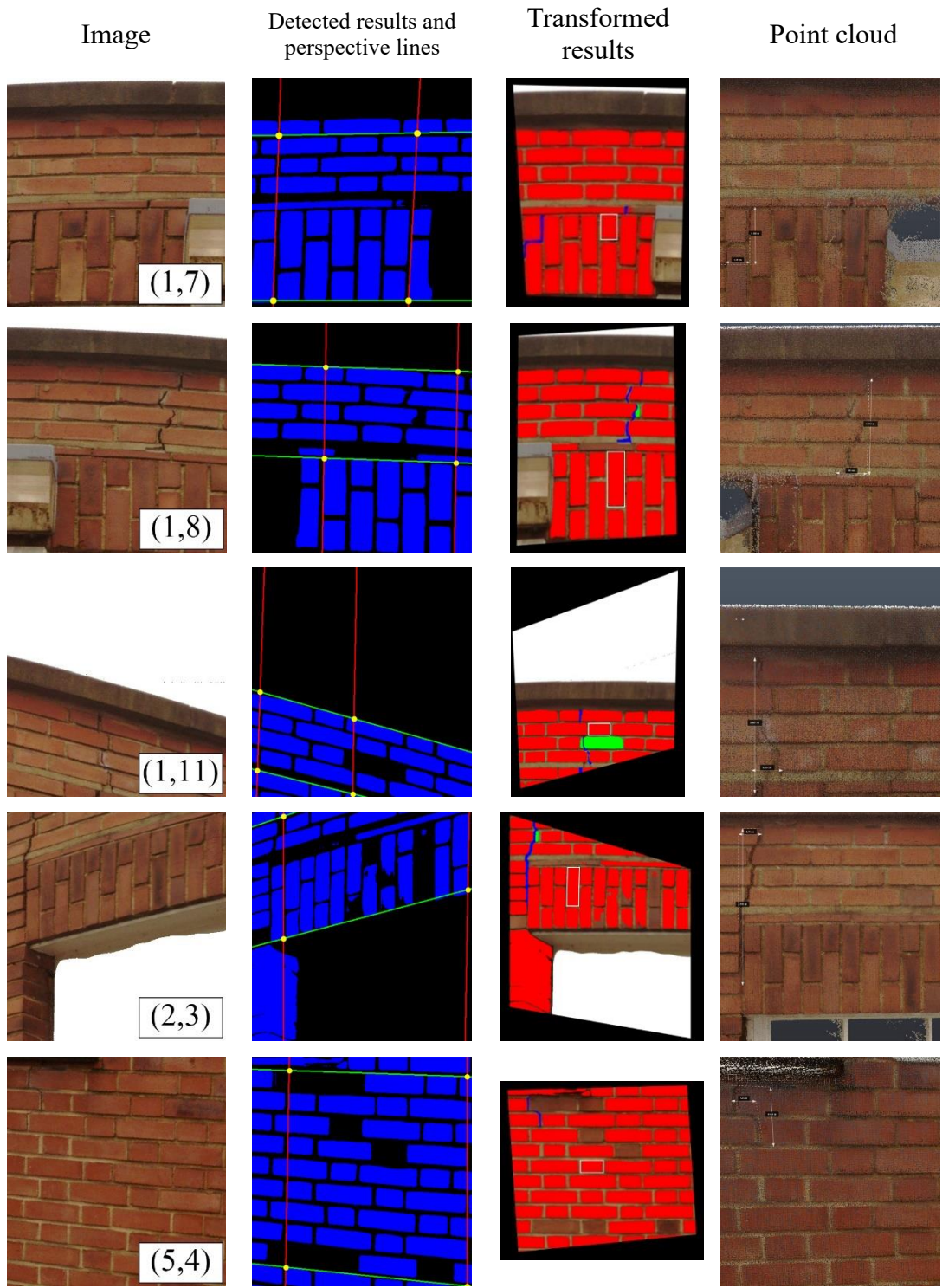

Figure 20. Example of crack and brick detection results display

The results in Figure 20 illustrate successful detection, capturing four main cracks along with a minor crack. Perspective lines were detected, and a perspective transformation has been applied. The transformed image's aspect ratio was corrected based on known brick unit dimensions, and the results are shown in the 'transformed results'. At this point, each transformed orthographic projection image allows for the determination of the real area per pixel. We compared overall length, width, and transverse width of the cracks with manually obtained results from the laser scanner. The results are presented in Table 10.

Table 10. Comparative analysis of crack size detection results between image-based and TLS-based methods.

| Crack | Method | Total width (mm) | Total Height (mm) | Max. transverse width (mm) |
|---|---|---|---|---|
| (1,7)* 1.76mm/pixel | Image-based | 79(45p) | 169(96) | 12(7) |
| | TLS-based | 78 | 185 | 11 |
| (1,8) 1.45mm/pixel | Image-based | 103(71p) | 286(196p) | 58(40p) |
| | TLS-based | 100 | 311 | 59 |
| ((1,11) 2.14mm/pixel | Image-based | 71 (33p) | 323(151p) | 19(9p) |
| | TLS-based | 82 | 361 | 17 |
| (2,3) 2.4mm/pixel | Image-based | 77(32p) | 518(216p) | 17(7p) |
| | TLS-based | 87 | 593 | 15 |
| (5,4) 1.62mm/pixel | Image-based | 68(42p) | 136(84p) | 11(7p) |
| | TLS-based | 58 | 151 | 11 |
| Mean Percentage Error | | 8.78 | 9.96 | 7.18 |

* Neglecting the small crack at the top right corner.

Here, the total width and total height are obtained by extracting the minimum coordinates ($x_{1min}$, $y_{1min}$) and maximum coordinates ($x_{2max}$, $y_{2max}$) from all detected bounding boxes ($x_1$, $y_1$, $x_2$, $y_2$) from all cracks in a single image, where $x_1$, $y_1$ represent the top-left corner of the bounding box while $x_2$, $y_2$ represent the bottom-right corner of the detection box. While the maximum transverse width of the crack refers to the maximum width of detected mask, along the x-axis direction, line by line. Mean Percentage Error (MPE) is calculated by the following equation:

$$MPE = \frac{100\%}{n} \sum_{t=1}^{n} \frac{a_t - f_t}{a_t} \qquad \text{Equation 8}$$

Where $a_t$ are the measured values from TLS, and $f_t$ represents our proposed monocular-based method. We can see from Table 10 the difference between TLS and the monocular-based method is generally within 10%. This indicates that the proposed method is effective. Even though we used panoramic image, where perspective lines are not straight, for the small cropped sub-images, deviations within a certain range are acceptable. Non-panoramic images may yield better results.

The examples from Figure 20 have already presented some cases of recognition failures. In the image at coordinates (2, 3), there are some vertically arranged bricks, which are underrepresented in the dataset, leading to recognition failure of several instance. On the other hand, at coordinates (5, 4), the number of instances exceeds the maximum query number (80) for the trained model, reaching the maximum matching quantity. As a result, additional instances cannot be recognized. Figure 21 shows some other examples of failed detection.

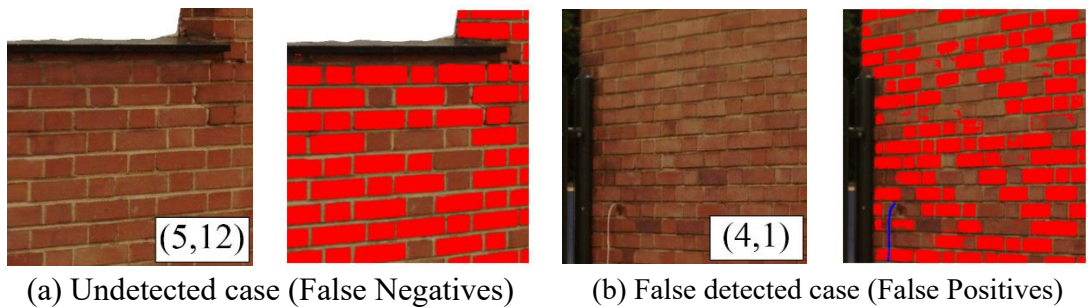

(a) Undetected case (False Negatives)   (b) False detected case (False Positives)

Figure 21. Examples of limitations in model detection.

In Figure 20(a), we see a tiny crack that was not detected. The main reasons are that the crack is very small, and the total predicted instances for the entire image have reached the limit of 80. This tiny crack, being a low-confidence instance, was not prioritized and failed to be matched, leading to recognition failure. Increasing the query number during model training or using smaller cropped sizes during inference can alleviate this issue. However, the former increases training time, while the latter extends inference time. The parameters for model training and inference can be tailored based on the specific application scenario and conditions. The second example shown in Figure 20(b) is a misidentification, where a cable is wrongly recognized as a crack. This is one of the common errors in deep learning. Expanding the dataset to allow the model to recognize more classes may help address such issues.

## 5. Conclusion

In this paper, we propose two novel instance segmentation algorithms based on the largest to-date fundamental segmentation model for the automation of crack detection tasks, with a specific emphasis on less-explored masonry structures. MCrack1300 dataset was established for model training. In the context of masonry, building upon the successful detection achieved by the proposed methods, a more reliable approach for estimating masonry crack dimensions is introduced. Through case studies, the effectiveness of the overall methodology is validated. The main contributions of this paper are:

- **A new dataset, MCrack1300, has been established to facilitate research in the field of masonry crack detection.** This dataset comprises various types of bricks along with cracks of diverse shapes and sizes, providing valuable resources for algorithm development and evaluation.
- **Two novel SAM-based methods were proposed to achieve automatic detection tasks in masonry, and both reach SOTA level.** It includes *Method I\**: LoRA-SAM with Mask2Former decoder and *Method II\**: LoRA-SAM with self-generating prompter. The two approaches above improve SAM's mask decoder and prompt encoder for automated execution, respectively. By compared with the previously best model, Mask2Former on MCrack1300 dataset, utilizing the official COCO instance segmentation evaluation methodology, *Method I\**, in terms of $AP_{mask}$, demonstrated an improvement of 3.6%, while *Method II\** exhibited an enhancement of 2.8%. Furthermore, SAM-based methods show advantages in segmenting objects with complex shapes. In the context of crack segmentation, two proposed methods demonstrated an increase of 6.1% in $AP_{mask}$, with respective increases of 7.3% (*Method I\**) and 9.5% (*Method II\**) in $AP_{mask}^{50}$, compared to the leading Mask2Former model.
- **Fine-tuning SAM's core components enables adaptation to downstream tasks without introducing specialized crack detection algorithm modules.** Two methods utilize the latest PEFT technique, LoRA, to fine-tune SAM heavy encoder, so as to adapt the downstream tasks. Using LoRA demonstrates an approximate 4% improvement shown in ablation study. While for *Method II\**, the mask decoder was unfrozen and brings around 8% improvement. The proposed methods do not incorporate any specific modules tailored for crack detection. They possess potential for further expansion into various engineering inspection tasks.

- **A simple five-stage feature pyramid structure has been proven to be effective for the non-hierarchical ViT.** In our instance segmentation tasks, using only the last layer from the non-hierarchical ViT and establishing a five-stage feature pyramid with two times up-sampling and two times down-sampling, and then transmit them to subsequent components, yielded reliable results with relatively low computational time.

- **A perspective transformation method based on the Hough Line Transform is proposed for accurately estimating the dimensions of masonry cracks.** Building upon the successful detection achieved by the previous proposed methods, this technique automatically extracts perspective lines on the masonry material background by excluding masks of broken bricks and cracks, and solely focusing on the boundaries of bricks. Consequently, this approach allowed for a more accurate calculation of the perspective matrix, and then transformed images into orthographic projection maps. With known size of the brick units, accurate estimations of crack dimensions are possible. Therefore, the proposed methods are capable of handling images from various perspectives, even under significant inclination angles. The results indicate that the difference between TLS and our method is generally within 10%.

There are some limitations to our research: Firstly, the dataset does not include masonry structures built with stone as the construction material. As for the proposed model, the training process is relatively slow, especially for self-generating prompter method. A large $N_p$ (greater than 50) will increase a lot in GPU memory requirements. This limitation arises because the mask decoder needs to perform a forward pass for each prompt received, and this is dependent on the number of instances in the image. Too few queries or $N_p$, when encountering a situation with too many image instances, can result in the inability to match all instances. Better self-generating prompter methods or improvements to the mask decoder might help to address these challenges. Moreover, the proposed method for estimating crack size based on brick units relies on perspective line detection and is applicable only to masonry structures with relatively uniform backgrounds.